%% file: main.tex
\documentclass[11pt,a4paper]{article}
\usepackage{naacl21}
\usepackage{times}
\usepackage{latexsym}

\usepackage[T1]{fontenc}

\usepackage[utf8]{inputenc}

\usepackage{hyperref}
\usepackage{url}
\usepackage{amsfonts}
\usepackage{xspace}
\usepackage{times}
\usepackage{latexsym}

\usepackage[ruled,vlined]{algorithm2e}
\usepackage{enumitem}
\usepackage{float}

\usepackage{microtype}
\usepackage{graphicx}
\usepackage{smile}
\usepackage{setspace}
\AtBeginDocument{%
  \addtolength\abovedisplayskip{-0.5\baselineskip}%
  \addtolength\belowdisplayskip{-0.5\baselineskip}%
  \addtolength\abovedisplayshortskip{-0.5\baselineskip}%
  \addtolength\belowdisplayshortskip{-0.5\baselineskip}%
}

\newcommand{\ours}{{\sf COSINE}}
\newcommand{\Init}{{\sf Init}}

\title{Fine-Tuning Pre-trained Language Model with Weak Supervision: \\ A Contrastive-Regularized Self-Training Approach}

\author{Yue Yu\thanks{Equal Contribution.} \ \ \ Simiao Zuo\footnotemark[1] \ \ \ Haoming Jiang \ \  Wendi Ren \ \  Tuo Zhao \ \   Chao Zhang \\
  Georgia Institute of Technology, Atlanta, GA, USA \\
  \texttt{\{yueyu,simiaozuo,jianghm,wren44,tourzhao,chaozhang\}@gatech.edu}\\}

\date{}

\begin{document}
\maketitle

\begin{abstract}
Fine-tuned pre-trained language models (LMs) have achieved enormous success in many natural language processing (NLP) tasks, but they still require excessive labeled data in the fine-tuning stage. We study the problem of fine-tuning pre-trained LMs using only weak supervision, without any labeled data. This problem is challenging because the high capacity of LMs makes them prone to overfitting the noisy labels generated by weak supervision.
To address this problem, we develop a contrastive self-training framework, {\ours}, to enable fine-tuning LMs with weak supervision. 
Underpinned by contrastive regularization and confidence-based reweighting,
our framework gradually improves model fitting while effectively suppressing error propagation.
Experiments on sequence, token, and sentence pair classification tasks show that our model outperforms the strongest baseline by large margins and achieves competitive performance with fully-supervised fine-tuning methods. Our implementation is available on \url{https://github.com/yueyu1030/COSINE}.
\end{abstract}

\input{1intro}

\input{2formulation}
\input{3method}
\input{5.0exp}
\input{5Relatedwork}
\input{6discussion}

\input{7conclusion}

\section*{Acknowledgments}
\vskip -0.05in
We thank anonymous reviewers for their feedbacks. This work was supported in part by the National
Science Foundation award III-2008334, Amazon
Faculty Award, and Google Faculty Award.

\bibliography{naacl2021}
\bibliographystyle{aclnatbib}

\appendix
{
\input{appendix_upload}
}

\end{document}

%% file: 1intro.tex
\vspace{-0.15in}
\section{Introduction}
\vspace{-0.1in}



Language model (LM)
pre-training and fine-tuning
achieve state-of-the-art performance in various natural language processing tasks~\cite{peters2018deep,devlin2019bert,liu2019roberta,raffel2019exploring}.
Such approaches stack task-specific layers on top of pre-trained
language models, \eg, BERT \cite{devlin2019bert},  then fine-tune the models with task-specific
data. During fine-tuning, the semantic and syntactic knowledge in the
pre-trained LMs is adapted for the target task.
Despite their success, one bottleneck for fine-tuning LMs is the requirement of labeled data. When labeled data are scarce, the fine-tuned models 
often suffer from degraded performance, and the large number of parameters 
can cause severe overfitting~\cite{uda}.

To relieve the label scarcity bottleneck,  we fine-tune the pre-trained language models with only weak supervision. 
While collecting large amounts of clean labeled data is expensive for many NLP tasks, it is often cheap to obtain weakly labeled data from various weak supervision sources, such as semantic rules \cite{Awasthi2020Learning}. For example, in sentiment analysis, we  can use rules \texttt{`terrible'}$\rightarrow$\texttt{Negative}
(a keyword rule) and \texttt{`* not recommend *'}$\rightarrow$\texttt{Negative}
(a pattern rule) to generate large amounts of weak labels.





Fine-tuning language models with weak supervision is nontrivial. Excessive label noise, \eg, wrong labels, and limited label coverage are common and inevitable in weak supervision. Although existing fine-tuning approaches~\cite{ensemble,Zhu2020FreeLB,smart} improve LMs' generalization ability, they are not designed for noisy data and are still easy to overfit on the noise. 
Moreover, existing works on tackling label noise are flawed and are not designed for fine-tuning LMs. For example,  \citet{ratner2019snorkel,varma2018snuba} use probabilistic models to aggregate multiple weak supervisions for denoising, but they 
generate weak-labels in a context-free manner, without using LMs to encode contextual information of the training samples~\cite{aina2019putting}.
Other works \cite{luo2017noise,wang2019learning} focus on noise transitions without explicitly conducting instance-level denoising, and they require clean training samples. Although some recent studies~\cite{Awasthi2020Learning,ren2020denoise} design labeling function-guided neural modules to denoise each sample, they require prior knowledge on weak supervision, which is often infeasible in real practice.

Self-training \cite{rosenberg2005semi,lee2013pseudo} is a proper tool for fine-tuning language models with weak supervision. It augments the training set with unlabeled data by generating pseudo-labels for them, which improves the models' generalization power. 
This resolves the limited coverage issue in weak supervision. However, one major challenge of self-training is that 
the algorithm still suffers from error propagation---wrong pseudo-labels can cause model performance to gradually deteriorate.


We propose a new algorithm {\ours}\footnote{Short for \underline{\textbf{Co}}ntrastive \underline{\textbf{S}}elf-Training for F\underline{\textbf{ine}}-Tuning Pre-trained Language Model.} that fine-tunes pre-trained LMs with only weak supervision.
{\ours} leverages both weakly labeled and unlabeled data, as well as suppresses label noise via contrastive self-training. Weakly-supervised learning enriches data with potentially noisy labels, and our contrastive self-training scheme fulfills the denoising purpose.
Specifically, contrastive self-training regularizes the feature space by pushing samples with the same pseudo-labels close while pulling samples with different pseudo-labels apart. Such regularization enforces representations of samples from different classes to be more distinguishable, such that the classifier can make better decisions. To suppress label noise propagation during contrastive self-training, we propose confidence-based sample reweighting and regularization methods. The reweighting strategy emphasizes samples with high prediction confidence, which are more likely to be correctly classified, in order to reduce the effect of wrong predictions. Confidence regularization encourages smoothness over model predictions, such that no prediction can be over-confident, and therefore reduces the influence of wrong pseudo-labels.

Our model is flexible and can be naturally extended to semi-supervised learning, where a small set of clean labels is available. Moreover, since we do not make assumptions about the nature of the weak labels, {\ours} can handle various types of label noise, including biased labels and randomly corrupted labels. Biased labels are usually generated by semantic rules, whereas corrupted labels are often produced by crowd-sourcing.



Our main contributions are: (1) A contrastive-regularized self-training framework that fine-tunes pre-trained LMs with only weak supervision.
(2) Confidence-based reweighting and regularization techniques that reduce error propagation and prevent over-confident predictions. 
(3) Extensive experiments on 6 NLP classification tasks using 7 public benchmarks verifying the efficacy of {\ours}. We highlight that our model achieves competitive performance in comparison with fully-supervised models on some datasets, \eg, on the Yelp dataset, we obtain a $97.2\%$ (fully-supervised) v.s. $96.0\%$ (ours) accuracy comparison. 

\begin{figure*}[!t]
  \centering
    \vspace{-0.1in}
    \includegraphics[width=14.1cm]{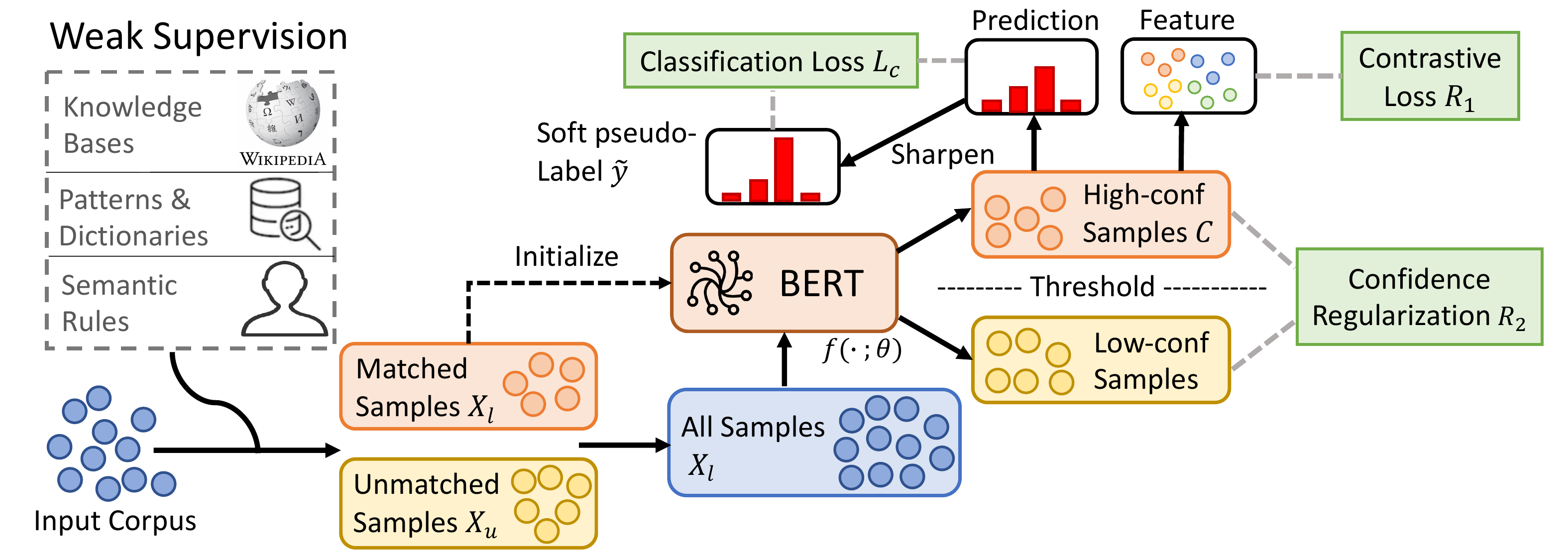}
    \caption{\textit{The framework of \ours. We first fine-tune the pre-trained language model on weakly-labeled data with early stopping. Then, we conduct contrastive-regularized self-training to improve model generalization and reduce the label noise. During self-training, we calculate the confidence of the prediction and update the model with high confidence samples to reduce error propagation.}
    }
    \vspace{-0.05in}
  \label{fig:framework}
\end{figure*}

%% file: 2formulation.tex
\vspace{-0.1in}
\section{Background}
\vspace{-0.1in}

In this section, we introduce weak supervision and our problem formulation.

\noindent \textbf{Weak Supervision.}
Instead of using human-annotated data, we obtain labels from weak supervision sources, including keywords and semantic rules\footnote{Examples of weak supervisions are in Appendix~\ref{app:LFs}.}. From weak supervision sources, each of the input samples $x \in \mathcal{X}$ is given a label $y \in \mathcal{Y} \cup \{\emptyset\}$, where $\mathcal{Y}$ is the label set and $\emptyset$ denotes the sample is not matched by any rules. For samples that are given multiple labels, \eg, matched by multiple rules, we determine their labels by majority voting. 

\noindent \textbf{Problem Formulation.} 
We focus on the weakly-supervised classification problems in natural language processing. We consider three types of tasks: sequence classification, token classification, and sentence pair classification. These tasks have a broad scope of applications in NLP, and some examples can be found in Table \ref{tab:task}.

Formally, the weakly-supervised classification problem is defined as the following: Given weakly-labeled samples $\mathcal{X}_l = \{(x_i,y_i)\}_{i=1}^L$ and unlabeled samples $\mathcal{X}_u = \{x_j\}_{j=1}^U$, we seek to learn a classifier $f(x;\theta): \mathcal{X} \rightarrow \mathcal{Y}$. Here $\mathcal{X}=\mathcal{X}_l \cup \mathcal{X}_u$ denotes all the samples and $\mathcal{Y}=\{1,2,\cdots,C\}$ is the label set, where $C$ is the number of classes.

\begin{table*}[t]
\small
	\begin{center}
		\begin{tabular}{cccc}
		   \toprule
		   \textbf{Formulation}  & \textbf{Example Task} & \textbf{Input} & \textbf{Output} \\
		   \midrule 
		   Sequence Classification  &
		   \begin{tabular}[c]{@{}c@{}}
		   Sentiment Analysis, Topic Classification, \\ Question Classification
		   \end{tabular}
		   & $\left[\bx_{1}, \dots, \bx_{N}\right]$ & $y$ \\ \hline
		   Token Classification  & 
		   \begin{tabular}[c]{@{}c@{}}
		   Slot Filling, Part-of-speech Tagging, \\ Event Detection
		   \end{tabular}
		   & $\left[x_{1}, \dots, x_{N}\right]$ & $\left[y_{1}, \dots, y_{N}\right]$\\ \hline
		   Sentence Pair Classification  & 
		   \begin{tabular}[c]{@{}c@{}}
		    Word Sense Disambiguation, Textual Entailment, \\  Reading Comprehension	
		   \end{tabular}
		    & $[\bx_1, \bx_2]$ & $y$  \\
		   \bottomrule
		\end{tabular}
	\end{center}
	\vspace{-0.1in}
	\caption{\textit{Comparison of different tasks. For sequence classification, input is a sequence of sentences, and we output a scalar label. For token classification, input is a sequence of tokens, and we output one scalar label for each token. For sentence pair classification, input is a pair of sentences, and we output a scalar label.} 
	}
	\vspace{-0.15in}
	\label{tab:task}
\end{table*}




%% file: 3method.tex
\vspace{-0.1in}
\section{Method}
\vspace{-0.1in}

Our classifier $f = g \circ \text{BERT}$ consists of two parts: $\text{BERT}$ is a pre-trained language model that outputs hidden representations of input samples, and $g$ is a task-specific classification head that outputs a $C$-dimensional vector, where each dimension corresponds to the prediction confidence of a specific class. In this paper, we use RoBERTa \cite{liu2019roberta} as the realization of $\text{BERT}$. 

The framework of {\ours} is shown in Figure~\ref{fig:framework}. First, {\ours} initializes the LM with weak labels. In this step, the semantic and syntactic knowledge of the pre-trained LM are transferred to our model. Then, it uses contrastive self-training to suppress label noise propagation and continue training.

\vspace{-0.05in}
\subsection{Overview}
\vspace{-0.05in}
\label{sec:pretrain}

The training procedure of \ours~is as follows.

\noindent \textbf{Initialization with Weakly-labeled Data.}
We fine-tune $f(\cdot; \theta)$ with weakly-labeled data $\mathcal{X}_l$ by solving the optimization problem
\begin{equation}
\min_{\theta}~\frac{1}{|\cX_l|}\sum_{(x_{i}, y_{i}) \in \cX_l} \text{CE}\left(f(x_{i}; \theta), y_{i}\right),
\label{eq:stage1}
\end{equation}
where $\text{CE}(\cdot, \cdot)$ is the cross entropy loss. We adopt early stopping~\cite{dodge2020fine} to prevent the model from overfitting to the label noise. However, early stopping causes underfitting, and we resolve this issue by contrastive self-training.

\noindent \textbf{Contrastive Self-training with All Data.} The goal of contrastive self-training is to leverage all data, both labeled and unlabeled, for fine-tuning, as well as to reduce the error propagation of wrongly labelled data. We generate pseudo-labels for the unlabeled data and incorporate them into the training set.
To reduce error propagation, we introduce contrastive representation learning (Sec.~\ref{sec:contrastive}) and confidence-based sample reweighting and regularization (Sec.~\ref{sec:denoise}).
We update the pseudo-labels (denoted by $\tilde{\by}$) and the model iteratively. The procedures are summarized in Algorithm \ref{alg:main}.

\noindent $\diamond$ \textbf{Update $\tilde{\by}$ with the current $\theta$.}  To generate the pseudo-label for each sample $x \in \mathcal{X}$, one straight-forward way is to use hard labels \cite{lee2013pseudo}
\begin{equation}
\tilde{y}_{\text{hard}}=\underset{j \in \mathcal{Y}}{\text{argmax}} \left[{f}(x; \theta)\right]_{j}.
\label{eq:hard}
\end{equation}
Notice that $f(x;\theta) \in \mathbb{R}^C$ is a probability vector and $[f(x;\theta)]_j$ indicates the $j$-th entry of it.
However, these hard pseudo-labels only keep the most likely class for each sample and result in the propagation of labeling mistakes.
For example, if a sample is mistakenly classified to a wrong class, assigning a 0/1 label complicates model updating (Eq.~\ref{eq:totalloss}), in that the model is fitted on erroneous labels.
To alleviate this issue, for each sample $x$ in a batch $\mathcal{B}$, we generate soft pseudo-labels\footnote{More discussions on hard vs.soft are in Sec.~\ref{sec:ablation}.} \cite{xieb16,uda,meng2020weakly,liang2020bond} $\tilde{\by} \in \mathbb{R}^C$ based on the current model as
\begin{equation}
\tilde{\by}_j= \frac{[{f}(x; \theta)]_{j}^2/f_j }{\sum_{j' \in \mathcal{Y}}[{f}(x; \theta)]_{j'}^2/f_{j'} },
\label{eq:reweilabel}
\end{equation}
where $f_j = \sum_{x' \in \mathcal{B}} [{f}(x';\theta)]^2_j$ is the sum over soft frequencies of class $j$.
The non-binary soft pseudo-labels guarantee that, even if our prediction is inaccurate, the error propagated to the model update step will be smaller than using hard pseudo-labels.

\noindent $\diamond$ \textbf{Update $\theta$ with the current $\tilde{\by}$.} 
We update the model parameters $\theta$ by minimizing
\begin{equation}
\cL(\theta; \tilde{\by})  = \cL_{c}(\theta; \tilde{\by}) + \cR_1(\theta; \tilde{\by}) + \lambda \cR_2(\theta),
\label{eq:totalloss}
\end{equation}
where $\mathcal{L}_c$ is the classification loss (Sec.~\ref{sec:denoise}), $\cR_1(\theta; \tilde{\by})$ is the contrastive regularizer (Sec.~\ref{sec:contrastive}), $\cR_2(\theta)$ is the confidence regularizer (Sec.~\ref{sec:denoise}), and $\lambda$ is the hyper-parameter for the regularization. 

\begin{algorithm}[t]
	\begin{small}
	\KwIn{Training samples $\mathcal{X}$; Weakly labeled samples $\mathcal{X}_l \subseteq \mathcal{X}$; Pre-trained LM $f(\cdot; \theta)$.}
	// \textit{Fine-tune the LM with weakly-labeled data.} \\
	\For{$t = 1, 2, \cdots, T_1$}{
	Sample a minibatch $\cB$ from $\mathcal{X}_l$.\\
	Update $\theta$ by Eq.~\ref{eq:stage1} using AdamW.
	}
	// \textit{Conduct contrastive self-training with all data.} \\
	\For{$t = 1, 2, \cdots, T_2$}{
		Update pseudo-labels $\tilde{\by}$ by Eq.~\ref{eq:reweilabel} for all $ x \in \mathcal{X}$. \\
		\For{$k = 1, 2, \cdots, T_3$}{
			Sample a minibatch $\cB$ from $\cX$. \\
			Select high confidence samples $\cC$ by Eq.~\ref{eq:C}. \\
            Calculate $\cL_c$ by Eq.~\ref{eq:newloss}, $\cR_1$ by Eq.~\ref{eq:contraloss}, $\cR_2$ by Eq.~\ref{eq:confreg}, and $\cL$ by Eq.~\ref{eq:totalloss}. \\
            Update $\theta$ using AdamW.
		}
	}
	\KwOut{Fine-tuned model $f(\cdot; \theta)$.}
	\end{small}
	\caption{\begin{small}Training Procedures of \ours. \end{small}}
	\label{alg:main}
\end{algorithm}

\input{4.3contrastive}

\vspace{-0.025in}
\subsection{Confidence-based Sample Reweighting and Regularization}
\label{sec:denoise}
\vspace{-0.05in}

While contrastive representations yield better decision boundaries, they require samples with high-quality pseudo-labels. In this section, we introduce reweighting and regularization methods to suppress error propagation and refine pseudo-label qualities.

\noindent \textbf{Sample Reweighting.}
In the classification task, samples with high prediction confidence are more likely to be classified correctly than those with low confidence. Therefore, we further reduce label noise propagation by a confidence-based sample reweighting scheme. For each sample $x$ with the soft pseudo-label $\tilde{\by}$, we assign $x$ with a weight $\omega(x)$ defined by
\begin{equation}
\omega = 1-\frac{H\left({\tilde{\by}}\right)}{\log (C)},~H(\tilde{\by}) = -\sum_{i=1}^{C}\tilde{\by}_i \log{\tilde{\by}_i},
\label{eq:rewconf}
\end{equation}
where $0 \leq H(\tilde{\by}) \leq \log(C)$ is the entropy of $\tilde{\by}$. Notice that if the prediction confidence is low, then $H(\tilde{\by})$ will be large, and the sample weight $\omega(x)$ will be small, and vice versa. We use a pre-defined threshold $\xi$ to select high confidence samples $\cC$ from each batch $\cB$ as 
\begin{equation}
\cC = \{x \in \cB~|~\omega(x) \geq \xi\}.
\label{eq:C}
\end{equation}
Then we define the loss function as
\begin{equation}
\mathcal{L}_c(\theta, \tilde{\by}) = \frac{1}{|\cC|} \sum_{x \in \cC} \omega(x) \cD_{\rm KL}\left(\tilde{\by} \| f(x;\theta) \right),
\label{eq:newloss}
\end{equation}
where
\begin{equation}
\mathcal{D}_{\rm KL}(P \| Q)=\sum_{k} p_{k} \log \frac{p_{k}}{q_{k}}
\label{eq:kldef}
\end{equation}
is the Kullback--Leibler (KL) divergence.


\noindent \textbf{Confidence regularization}
The sample reweighting approach promotes high confidence samples during contrastive self-training. However, this strategy relies on wrongly-labeled samples to have low confidence, which may not be true unless we prevent over-confident predictions. To this end, we propose a confidence-based regularizer that encourages smoothness over predictions, defined as
\begin{equation}
\mathcal{R}_2(\theta) =  \frac{1}{|\cC|} \sum_{x \in \cC} \mathcal{D}_{\text{KL}}\left(\mathbf{u} \| {f}(x; \theta)\right),
\label{eq:confreg}
\end{equation}
where $\mathcal{D}_{\text{KL}}$ is the KL-divergence and $\mathbf{u}_i=1/C$ for $i=1, 2, \cdots, C$. Such term constitutes a regularization to prevent over-confident predictions and leads to better generalization~\cite{labelsmoothing}.

%% file: 4.3contrastive.tex
\vspace{-0.05in}
\subsection{Contrastive Learning on Sample Pairs}
\label{sec:contrastive}
\vspace{-0.05in}

The key ingredient of our contrastive self-training method is to learn representations that encourage data within the same class to have similar representations and keep data in different classes separated. Specifically, we first select high-confidence samples (Sec.~\ref{sec:denoise}) $\cC$ from $\cX$. Then for each pair $x_i, x_j \in \cC$, we define their similarity as
\begin{equation}
W_{i j}=\left\{\begin{array}{ll}
1, & \text { if } \underset{k \in \cY}{\text{argmax}}[\tilde{\by}_i]_k = \underset{k \in \cY}{\text{argmax}}[\tilde{\by}_j]_k \\
0, & \text { otherwise }
\end{array}\right. ,
\label{eq:edgeweight}
\end{equation}
where $\tilde{\by}_i$, $\tilde{\by}_j$ are the soft pseudo-labels (Eq.~\ref{eq:reweilabel}) for $x_i$, $x_j$, respectively. For each $x \in \cC$, we calculate its representation $\bv=\text{BERT}(x)\in \mathbb{R}^{d}$, then we define the contrastive regularizer as
\begin{equation}
\cR_1(\theta; \tilde{\by})=\sum_{(x_{i}, x_{j}) \in \mathcal{C}\times \mathcal{C}} \ell(\bv_{i}, \bv_{j}, W_{i j}),
\label{eq:contraloss}
\end{equation}
where
\begin{equation}
\ell = W_{ij}d_{ij}^2+ (1 - W_{ij})[\max(0, \gamma-d_{ij})]^2.
\label{eq:ell}
\end{equation}
Here, $\ell(\cdot,\cdot,\cdot)$ is the contrastive loss \cite{chopra2005learning,Taigman_2014_CVPR}, $d_{ij}$ is the distance\footnote{We use scaled Euclidean distance $d_{i j}=\frac{1}{d}\left\|\boldsymbol{v}_{i}-\boldsymbol{v}_{j}\right\|_{2}^{2}$ by default. More discussions on $W_{ij}$ and $d_{ij}$ are in Appendix~\ref{appendix:distance}.} between $\bv_i$ and $\bv_j$, and $\gamma$ is a pre-defined margin.

\begin{figure}[t]
  \centering
    \includegraphics[width=7.6cm]{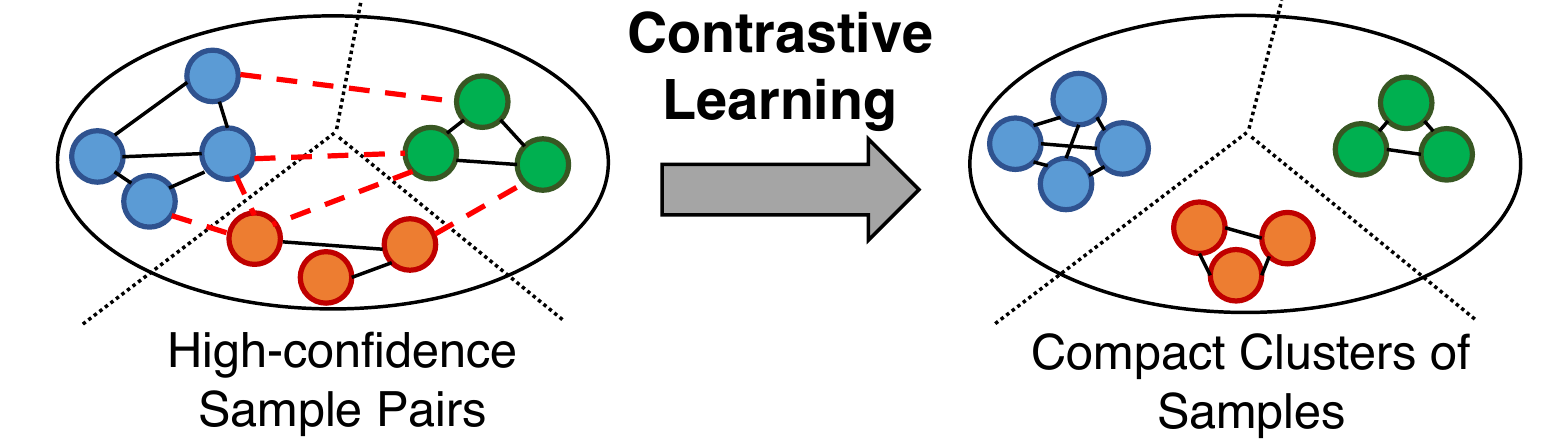}
    \vspace{-0.05in}
    \caption{\textit{An illustration of contrastive learning. The black solid lines indicate similar sample pairs, and the red dashed lines indicate dissimilar pairs.}}
   \vspace{-0.1in}
  \label{fig:contrastive}
\end{figure}

For samples from the same class, \ie $W_{ij}=1$,
Eq.~\ref{eq:contraloss} penalizes the distance between them, and for samples from different classes, the contrastive loss is large if their distance is small. In this way, the regularizer enforces similar samples to be close, while keeping dissimilar samples apart by at least $\gamma$.
Figure~\ref{fig:contrastive} illustrates the contrastive representations. We can see that our method produces clear inter-class boundaries and small intra-class distances, which  eases the classification tasks.


%% file: 5.0exp.tex
\vspace{-0.1in}
\section{Experiments}
\vspace{-0.05in}
\textbf{Datasets and Tasks.} We conduct experiments on 6 NLP classification tasks using 7 public benchmarks: 
\textit{AGNews}~\cite{AGNews} is a Topic Classification task;
\textit{IMDB}~\cite{IMDB} and \textit{Yelp}~\cite{meng2018weakly} are Sentiment Analysis tasks; \textit{TREC}~\cite{TREC} is a Question Classification task;
\textit{MIT-R}~\cite{mitr} is a Slot Filling task;
\textit{Chemprot}~\cite{chemprot} is a Relation Classification task;
and \textit{WiC}~\cite{pilehvar2019wic} is a Word Sense Disambiguation (WSD) task.
The dataset statistics are summarized in Table \ref{tab:dataset}. More details on datasets and weak supervision sources are in Appendix~\ref{app:dataset}\footnote{Note that we use the same weak supervision signals/rules for both our method and all the baselines for fair comparison.}.

\begin{table*}[tb!]
\centering
\begin{small}
		\begin{tabular}{ cccccccc}
			\toprule \bf Dataset & \bf Task & \bf Class &\bf \# Train & \bf \# Dev & \bf \# Test & \bf Cover & \bf Accuracy \\ \midrule
            AGNews &Topic &4 &96k & 12k & 12k & 56.4 & 83.1 \\ 
            IMDB &Sentiment &2 &20k & 2.5k & 2.5k & 87.5 & 74.5 \\ 
            Yelp &Sentiment& 2&30.4k & 3.8k & 3.8k & 82.8 & 71.5 \\ 
            MIT-R & Slot Filling& 9& 6.6k & 1.0k & 1.5k & 13.5 & 80.7 \\ 
            TREC & Question & 6& 4.8k & 0.6k & 0.6k & 95.0  & 63.8 \\ 
            Chemprot & Relation & 10 & 12.6k & 1.6k & 1.6k &  85.9 & 46.5  \\
            WiC & WSD & 2 & 5.4k & 0.6k & 1.4k & 63.4 & 58.8 \\
            \bottomrule
		\end{tabular}
	\caption{\textit{Dataset statistics. Here cover (in \%) is the fraction of instances covered by weak supervision sources in the training set, and accuracy (in \%) is the precision of weak supervision.}
	}
	\label{tab:dataset}
\end{small}
\vskip -0.05in
\end{table*}


\noindent \textbf{Baselines.} 
We compare our model with different groups of baseline methods:

\noindent (i) \textbf{Exact Matching~(ExMatch)}: The test set is directly labeled by weak supervision sources.


\noindent (ii) \textbf{Fine-tuning Methods}: The second group of baselines are fine-tuning methods for LMs:

\noindent $\diamond$\textit{RoBERTa}~\cite{liu2019roberta} uses the RoBERTa-base model with task-specific classification heads.

\noindent $\diamond$~\textit{Self-ensemble}~\citep{ensemble} uses 
self-ensemble and distillation to improve  performances. 

\noindent $\diamond$ \textit{FreeLB}~\cite{Zhu2020FreeLB} adopts adversarial training to enforce smooth outputs.

\noindent $\diamond$ \textit{Mixup}~\cite{zhang2018mixup} 
creates virtual training samples by linear interpolations. 

\noindent $\diamond$ \textit{SMART}~\cite{smart}  adds adversarial and smoothness constraints to fine-tune LMs and achieves state-of-the-art result for many NLP tasks.

\noindent (iii) \textbf{Weakly-supervised Models}: The third group of baselines are weakly-supervised models\footnote{All methods use RoBERTa-base as the backbone unless otherwise specified.}:

\noindent $\diamond$ \textit{Snorkel}~\cite{ratner2019snorkel} aggregates different labeling functions based on their correlations.

\noindent $\diamond$ \textit{WeSTClass}~\cite{meng2018weakly} trains a classifier with generated pseudo-documents and use self-training to bootstrap over all samples. 

\noindent $\diamond$ \textit{ImplyLoss}~\cite{Awasthi2020Learning} co-trains a rule-based classifier and a neural classifier to denoise.

\noindent $\diamond$ \textit{Denoise}~\cite{ren2020denoise} uses attention network to estimate reliability of weak supervisions, and then reduces the noise by aggregating weak labels.

\noindent $\diamond$ \textit{UST}~\cite{mukherjee2020uncertainty} is state-of-the-art for self-training with limited labels. It estimates  uncertainties via MC-dropout~\cite{mcdropout}, and then select samples with low  uncertainties for self-training.

\noindent \textbf{Evaluation Metrics.} We use classification accuracy on the test set as the evaluation metric for all datasets except MIT-R. MIT-R contains a large number of tokens that are labeled as ``Others''. We use the micro $F_1$ score from other classes for this dataset.\footnote{The Chemprot dataset also contains ``Others'' type, but such instances are few, so we still use accuracy as the metric.}


\noindent \textbf{Auxiliary.}
We implement {\ours} using PyTorch\footnote{\url{https://pytorch.org/}}, and we use RoBERTa-base as the pre-trained LM. 
Datasets and weak supervision details are in Appendix~\ref{app:dataset}. Baseline settings are in Appendices~\ref{app:baseline-settings}. Training details and setups are in Appendix~\ref{appendix:exp}. Discussions on early-stopping are in Appendix~\ref{appendix:early}. Comparison of distance metrics and similarity measures are in Appendix~\ref{appendix:distance}.

\input{5.1main}

\input{5.5flip}
\input{5.4semi}

\input{5.7supervised}

\input{5.6case}

\input{5.2ablation}

%% file: 5.1main.tex
\subsection{Learning From Weak Labels}
\label{sec:mainresult}
\begin{table*}[htb!]
	\begin{small}
	\centering
		\resizebox{1.99\columnwidth}{!}{
			\begin{tabular}{lccccccc}
				\toprule
				\textbf{Method} & \  \bf AGNews \ &  \bf \ IMDB \ & \ \bf Yelp \ &  \ \bf MIT-R \  &  \ \bf TREC  \ &  \ \bf Chemprot \  &  \ \bf WiC (dev)\ \\
				\midrule
				ExMatch  & $52.31$ & $71.28$ & $68.68$ & $34.93$  &$60.80$ & $46.52$ & $58.80$\\
				\hline 
				\multicolumn{6}{l}{\textbf{Fully-supervised Result}}\\
				RoBERTa-CL$^\diamond$~\cite{liu2019roberta} & $91.41$ & $94.26$& $97.27$ & $88.51$ &$96.68$ & $79.65$  & $70.53$\\
				\hline 
				\multicolumn{6}{l}{\textbf{Baselines}}\\
				RoBERTa-WL$^\dagger$~\cite{liu2019roberta} &$82.25$&$72.60$ & $74.89$ & $70.95$  & $62.25$ & $44.80$ & $59.36$\\
				Self-ensemble~\cite{ensemble} & $85.72$ & $86.72$ & $80.08$ & $72.88$ & $66.18$ & $44.62$  &  $62.71$\\ 
				FreeLB~\cite{Zhu2020FreeLB} & $85.12$ & $88.04$ & $85.68$ & $73.04$  & $67.33$ & $45.68$ & $63.45$\\
				Mixup~\cite{zhang2018mixup} & $85.40$ & $86.92$ & $92.05$ & $73.68$ & $66.83$ & $51.59$ &  $64.88$ \\
				SMART~\cite{smart} & $86.12$ & $86.98$ & $88.58$ & $73.66$  & $68.17$ & $48.26$ & $63.55$ \\ 
				\hline
				Snorkel~\cite{ratner2019snorkel} & $62.91$ & $73.22$ & $69.21$ & $ 20.63$ & $58.60$ & $37.50$ & -\!-\!-$^\ast$ \\
				WeSTClass~\cite{meng2018weakly} & $82.78$ &  $77.40$ & $76.86$ & -\!-\!-$^\otimes$ & $37.31$ & -\!-\!-$^\otimes$ & $48.59$\\
				ImplyLoss~\cite{Awasthi2020Learning} &$68.50$ & $63.85$ & $76.29$ & $74.30$ & $80.20$ & ${53.48}$ & $54.48$\\
				Denoise~\cite{ren2020denoise} & $85.71$ & $82.90$ & $87.53$ & $70.58$ & $69.20$ & $50.56$ & $62.38$ \\ 
				UST~\cite{mukherjee2020uncertainty} & $86.28$ & $84.56$ & $90.53$& $74.41$ & $65.52$ & $52.14$ & $63.48$ \\
				\hline 
				\hline
				\multicolumn{6}{l}{\textbf{Our} \ours \ \textbf{Framework}}\\
				\Init & $84.63$ & $83.58$ & $81.76$ & $72.97$ &$65.67$ & $51.34$ & $63.46$\\
			\ours & $\textbf{87.52}$ & $\textbf{90.54}$ & $\textbf{95.97}$& $\textbf{76.61}$ & $\textbf{82.59}$ & $\textbf{54.36}$ & $\textbf{67.71}$\\
				\bottomrule
			\end{tabular}
		}%

	$^\diamond$: RoBERTa is trained with clean labels.  $^\dagger$: RoBERTa is trained with weak labels. $^\ast$: unfair comparison. $^\otimes$: not applicable.
	\vspace{-0.05in}
	\caption{\textit{Classification accuracy (in \%) on various datasets. We report the mean over three runs.}}
	\label{tb:main_result}
	\end{small}
	\vspace{-0.05in}
\end{table*}

We summarize the weakly-supervised leaning results in Table~\ref{tb:main_result}. In all the datasets, \ours ~outperforms all the baseline models. A special case is the WiC dataset, where we use WordNet\footnote{\url{https://wordnet.princeton.edu/}} to generate weak labels. However, this enables Snorkel to access some labeled data in the development set, making it unfair to compete against other methods. We will discuss more about this dataset in Sec.~\ref{sec:semi}.

In comparison with directly fine-tuning the pre-trained LMs with weakly-labeled data, our model employs an ``earlier stopping'' technique\footnote{We discuss this technique in Appendix \ref{appendix:early}.} so that it does not overfit on the label noise. As shown, indeed ``\Init'' achieves better performance, and it serves as a good initialization for our framework.
Other fine-tuning methods and weakly-supervised models either cannot harness the power of pre-trained language models, \eg, Snorkel, or rely on clean labels, \eg, other baselines.
We highlight that although UST, the state-of-the-art method to date, achieves strong performance under few-shot settings, their approach cannot estimate confidence well with noisy labels, and this yields inferior performance.
Our model can gradually correct wrong pseudo-labels and mitigate error propagation via contrastive self-training. 

It is worth noticing that on some datasets, \eg, AGNews, IMDB, Yelp, and WiC, our model achieves the same level of performance with models (RoBERTa-CL) trained with clean labels. This makes {\ours} appealing in the scenario where only weak supervision is available.

%% file: 5.5flip.tex
\begin{figure}[t]
	\centering
	\vspace{-0.1in}
	\includegraphics[width=0.45\textwidth]{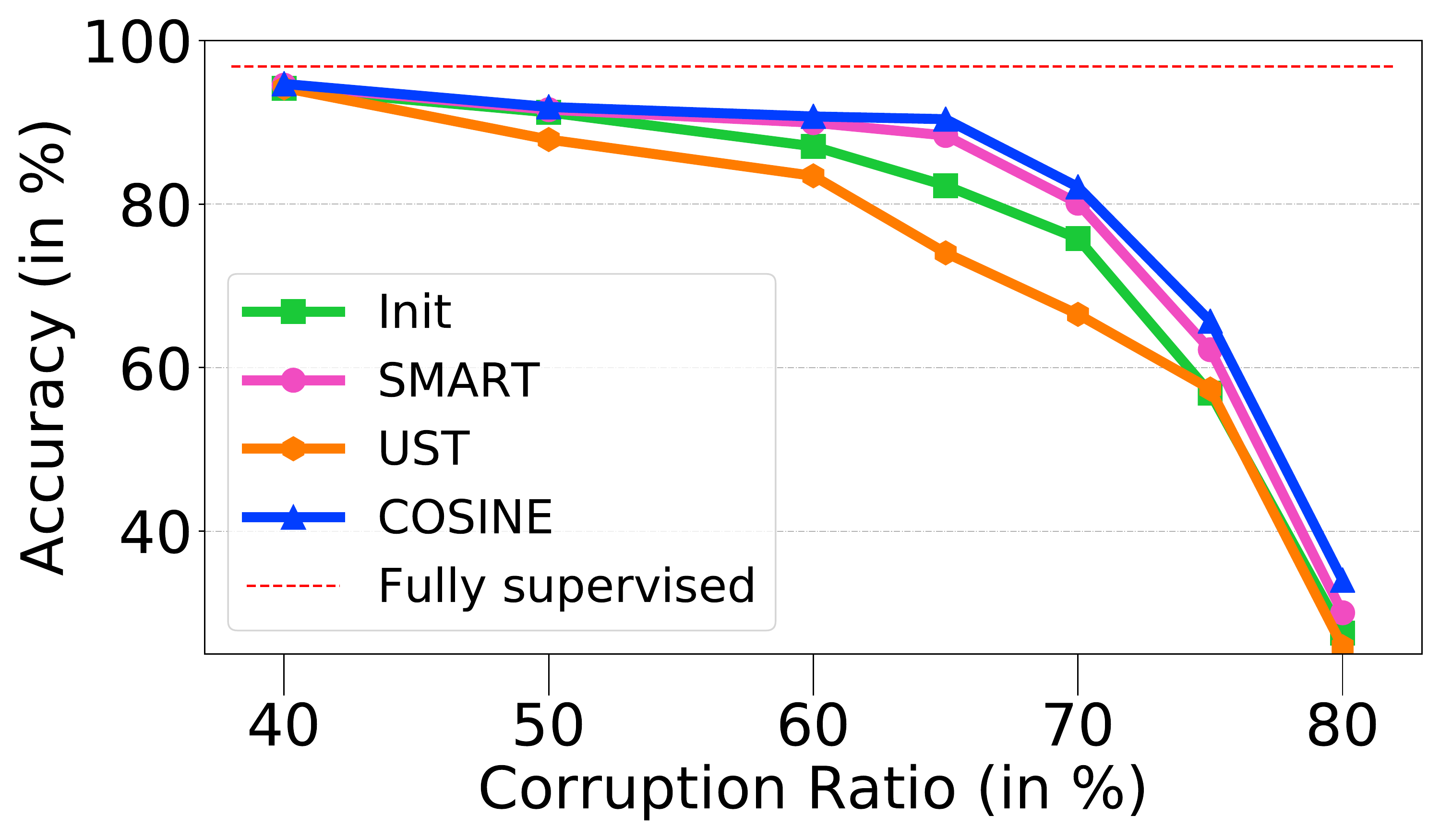}
	\vspace{-0.1in}
	\caption{\textit{Results of label corruption on TREC. When the corruption ratio is less than 40\%, the performance is close to the fully supervised method.}}
	\label{fig:flip}
	\vspace{-0.25in}
\end{figure}


\vspace{-0.1in}
\subsection{Robustness Against Label Noise}
\vspace{-0.05in}
Our model is robust against excessive label noise. We corrupt certain percentage of labels by randomly changing each one of them to another class. This is a common scenario in crowd-sourcing, where we assume human annotators mis-label each sample with the same probability. Figure~\ref{fig:flip} summarizes experiment results on the TREC dataset. Compared with advanced fine-tuning and self-training methods (\eg SMART and UST)\footnote{Note that some methods in Table~\ref{tb:main_result}, \eg, ImplyLoss and Denoise, are not applicable to this setting since they require weak supervision sources, but none exists in this setting.},
our model consistently outperforms the baselines.

%% file: 5.4semi.tex
\begin{table}[tb!]
\small
 	\vspace{-0.1in}
    \resizebox{1.03\columnwidth}{!}{
		\begin{tabular}{l@{\hskip3pt}|@{\hskip3pt} c@{\hskip3pt}|@{\hskip3pt} c@{\hskip3pt}|c}
		   \toprule
		   Model  & Dev  & Test  & \#Params \\
		   \hline 
		   Human Baseline & \multicolumn{2}{c|}{80.0}  & -\!-\!- \\
		   \hline
		   BERT~\cite{devlin2019bert}  & -\!-\!- & 69.6 & 335M\\
		   RoBERTa~\citep{liu2019roberta} & 70.5 & 69.9 & 356M \\
		   T5~\citep{raffel2019exploring} & -\!-\!- & 76.9 & 11,000M\\
		   \hline 
		   \hline 
		   \multicolumn{4}{l}{\textbf{Semi-Supervised Learning}}\\
		   \hline 
		   SenseBERT~\citep{levine2019sensebert} & -\!-\!- & 72.1 & 370M\\
		   RoBERTa-WL$^\dagger$~\citep{liu2019roberta}  & 72.3 & 70.2 & 125M\\
		   w/ MT$^\dagger$~\citep{tarvainen2017mean}  & 73.5 & 70.9 & 125M\\
		   w/ VAT$^\dagger$~\citep{miyato2018virtual}  & 74.2 & 71.2 & 125M\\
		   w/ {\ours}$^\dagger$  & \textbf{76.0} & \textbf{73.2} & 125M\\
		   \hline 
		   \hline 
		   \multicolumn{4}{l}{\textbf{Transductive Learning}}\\
		   \hline 
		   Snorkel$^\dagger$~\citep{ratner2019snorkel} & 80.5 & -\!-\!- & 1M \\
		   RoBERTa-WL$^\dagger$~\citep{liu2019roberta}  & 81.3 & 76.8 & 125M\\
		   w/ MT$^\dagger$~\citep{tarvainen2017mean}  & 82.1 & 77.1 & 125M\\
		   w/ VAT$^\dagger$~\citep{miyato2018virtual}  & 84.9 & 79.5 & 125M\\
		   w/ {\ours}$^\dagger$  & \textbf{89.5} & \textbf{85.3} & 125M\\
		   \bottomrule
		\end{tabular}
	\vspace{-0.15in}}
	\vskip -0.1in
	\caption{\textit{Semi-supervised Learning on WiC. VAT (Virtual Adversarial Training) and MT (Mean Teacher) are semi-supervised methods. $^\dagger$: has access to weak labels.}}
	\label{tab:semi-wic}
	\vspace{-0.25in}
\end{table}
	
	\vspace{-0.1in}
\subsection{Semi-supervised Learning}
	\vspace{-0.05in}
\label{sec:semi}
We can naturally extend our model to semi-supervised learning, where clean labels are available for a portion of the data. We conduct experiments on the WiC dataset. As a part of the SuperGLUE \cite{wang2019superglue} benchmark, this dataset proposes a challenging task: models need to determine whether the same word in different sentences has the same sense (meaning).

Different from previous tasks where the labels in the training set are noisy, in this part, we utilize the clean labels provided by the WiC dataset. We further augment the original training data of WiC with unlabeled sentence pairs obtained from lexical databases (\eg, WordNet, Wictionary). {Note that part of the unlabeled data can be weakly-labeled by rule matching.} This essentially creates a \textit{semi-supervised} task, where we have labeled data, weakly-labeled data and unlabeled data.

Since the weak labels of WiC are generated by WordNet and partially reveal the true label information, Snorkel \citep{ratner2019snorkel} takes this unfair advantage by accessing the unlabeled sentences and weak labels of validation and test data. To make a fair comparison to Snorkel, we consider the transductive learning setting, where we are allowed access to the same information by integrating unlabeled validation and test data and their weak labels into the training set. 
As shown in Table \ref{tab:semi-wic}, {\ours} with transductive learning achieves better performance compared with Snorkel.  Moreover, in comparison with semi-supervised baselines (\ie VAT and MT) and fine-tuning methods with extra resources (\ie, SenseBERT), {\ours} achieves better performance in both semi-supervised and transductive learning settings.



%% file: 5.7supervised.tex
\begin{figure*}[t!]
    \vspace{-0.38in}
        \centering
        \hspace{-3mm}
        \subfigure{
            \includegraphics[width=0.248\textwidth]{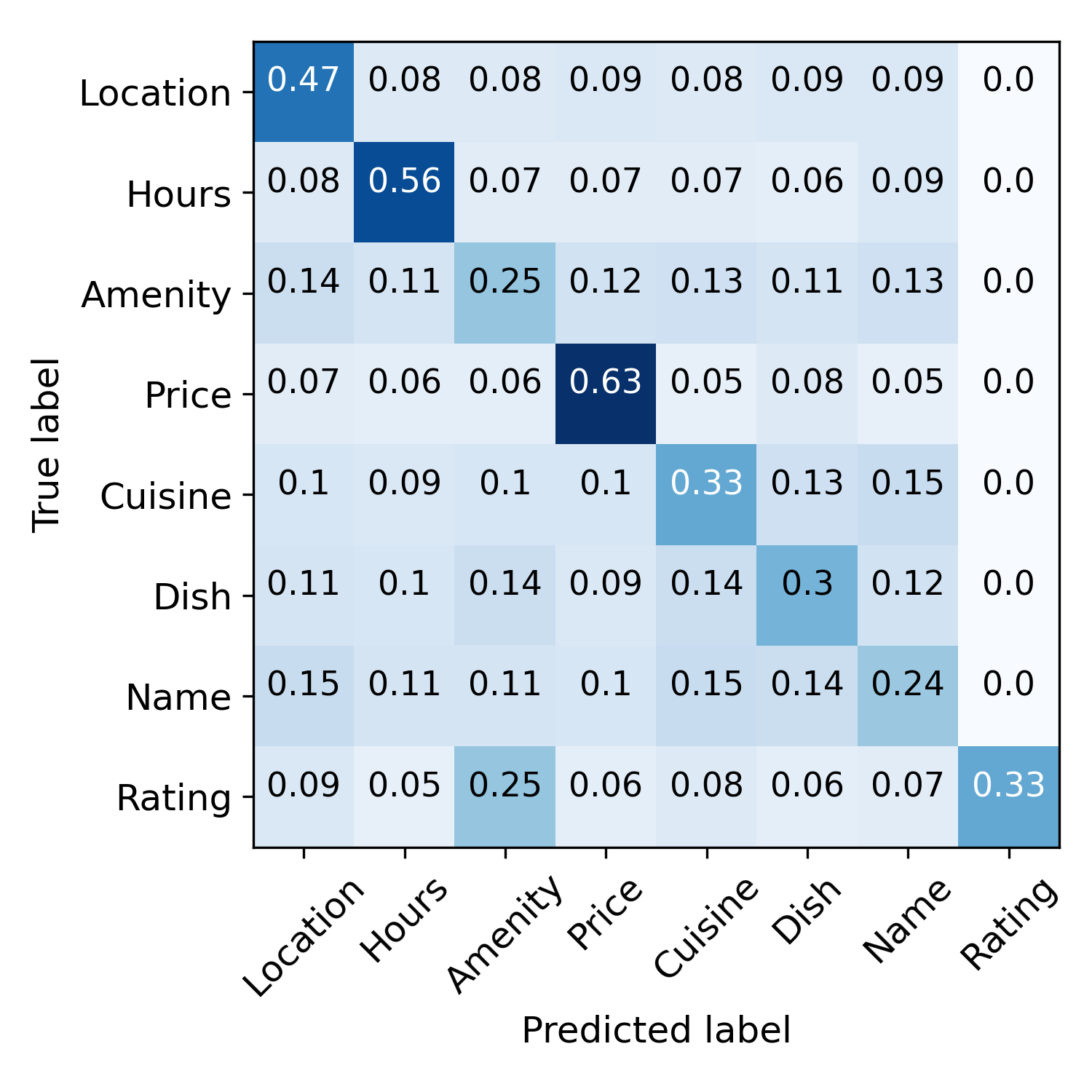}
            \vspace{-0.2in}
        }\hspace{-2.5mm}
        \subfigure{
            \includegraphics[width=0.248\textwidth]{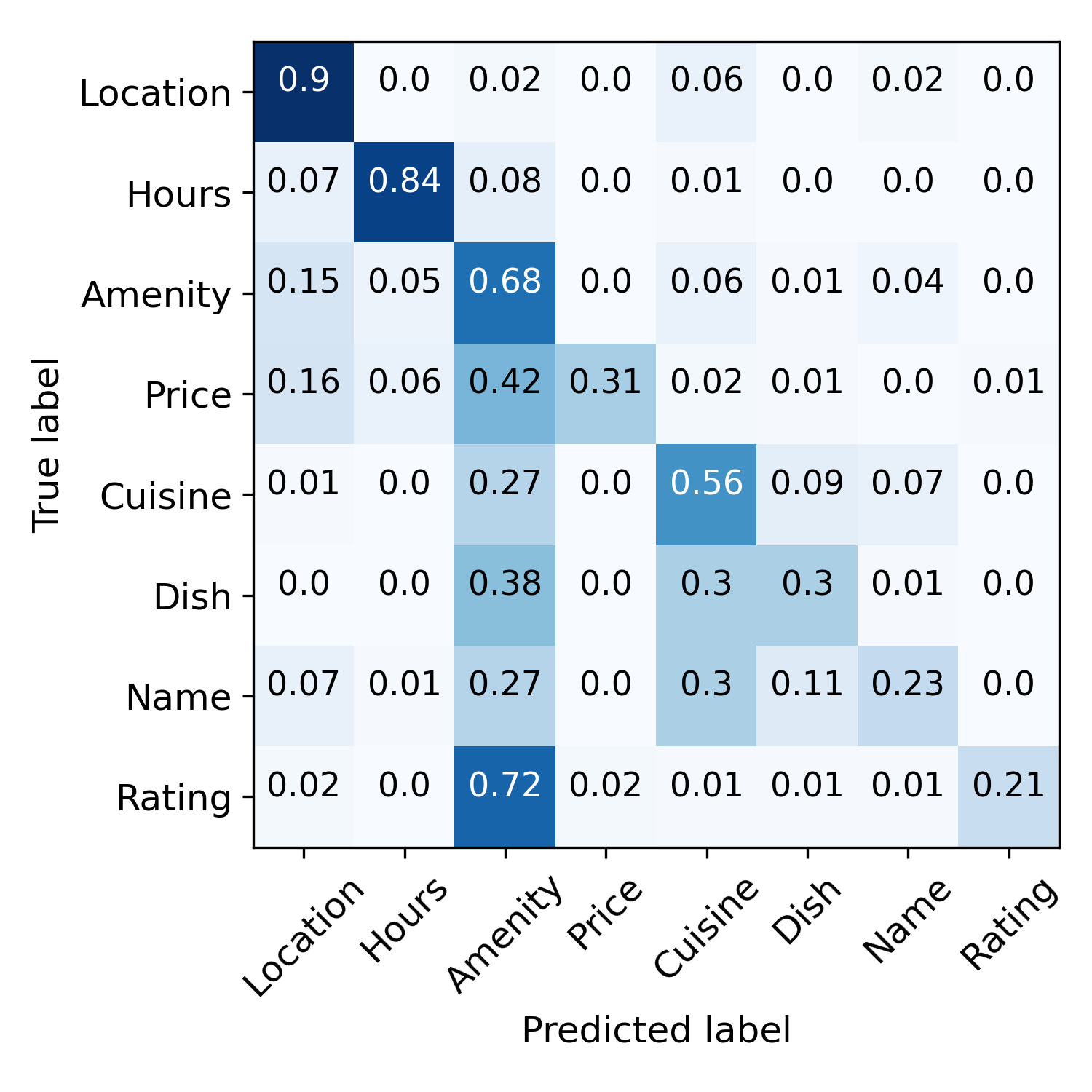}
            \vspace{-0.2in}
        }\hspace{-2.5mm}
         \subfigure{
            \includegraphics[width=0.248\textwidth]{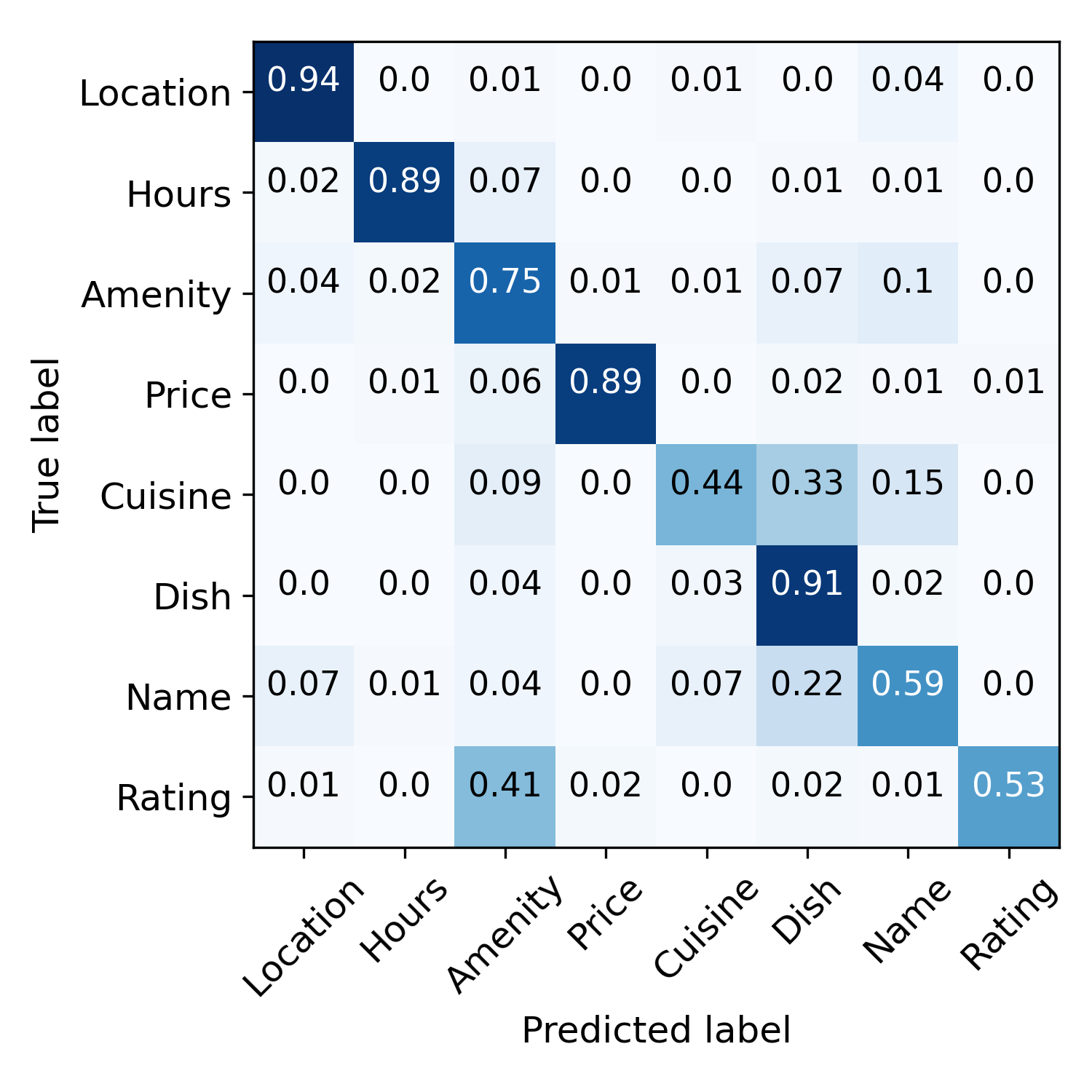}
            \vspace{-0.2in}
        }
        \hspace{-2.5mm}
       \subfigure{ \includegraphics[width=0.24\textwidth]{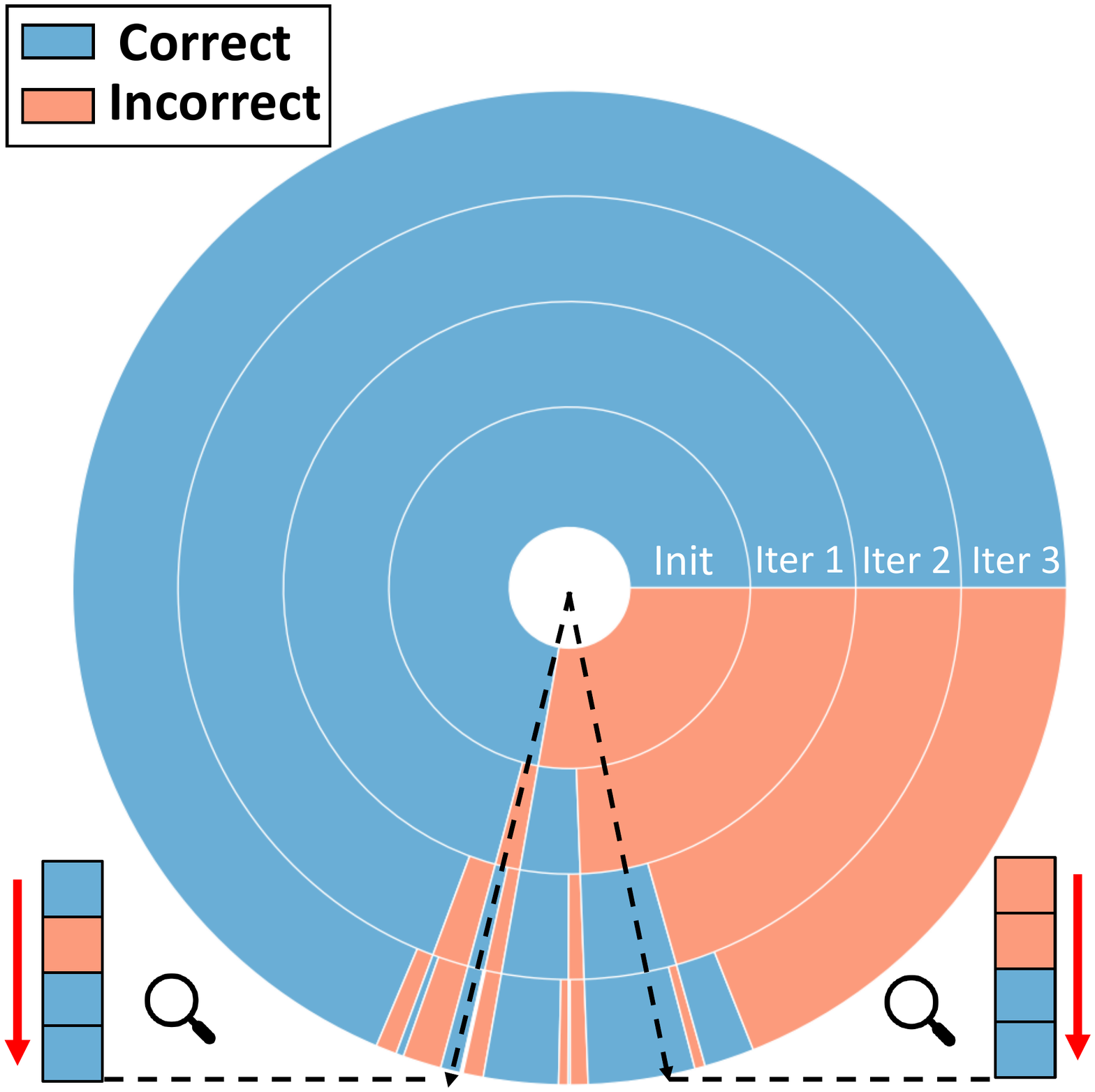}
\vspace{-0.25in}}
\vspace{-0.1in}
\caption{\textit{Classification performance on MIT-R. From left to right: visualization of ExMatch, results after the initialization step, results after contrastive self-training, and wrong-label correction during self-training.}}
        \label{fig:case}
\end{figure*}

\begin{figure*}
    \centering
    \begin{minipage}{.75\textwidth}
        \centering
        \subfigure[\textit{Effect of $\xi$.}]{
            \includegraphics[width=0.32\textwidth]{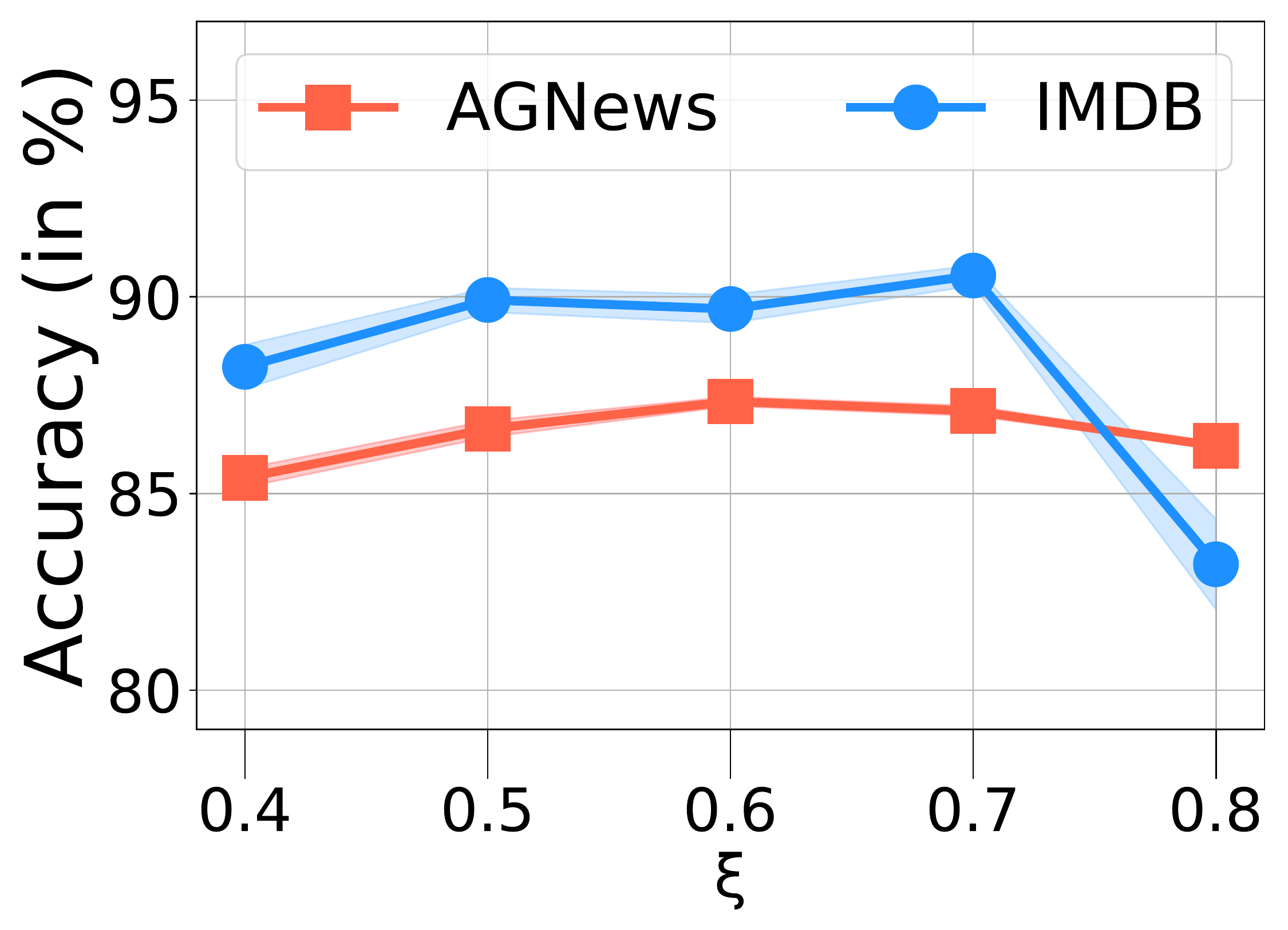}
            \label{fig:xi}
        }\hspace{-2.mm}
        \subfigure[\textit{Effect of $T_1$.}]{
            \includegraphics[width=0.32\textwidth]{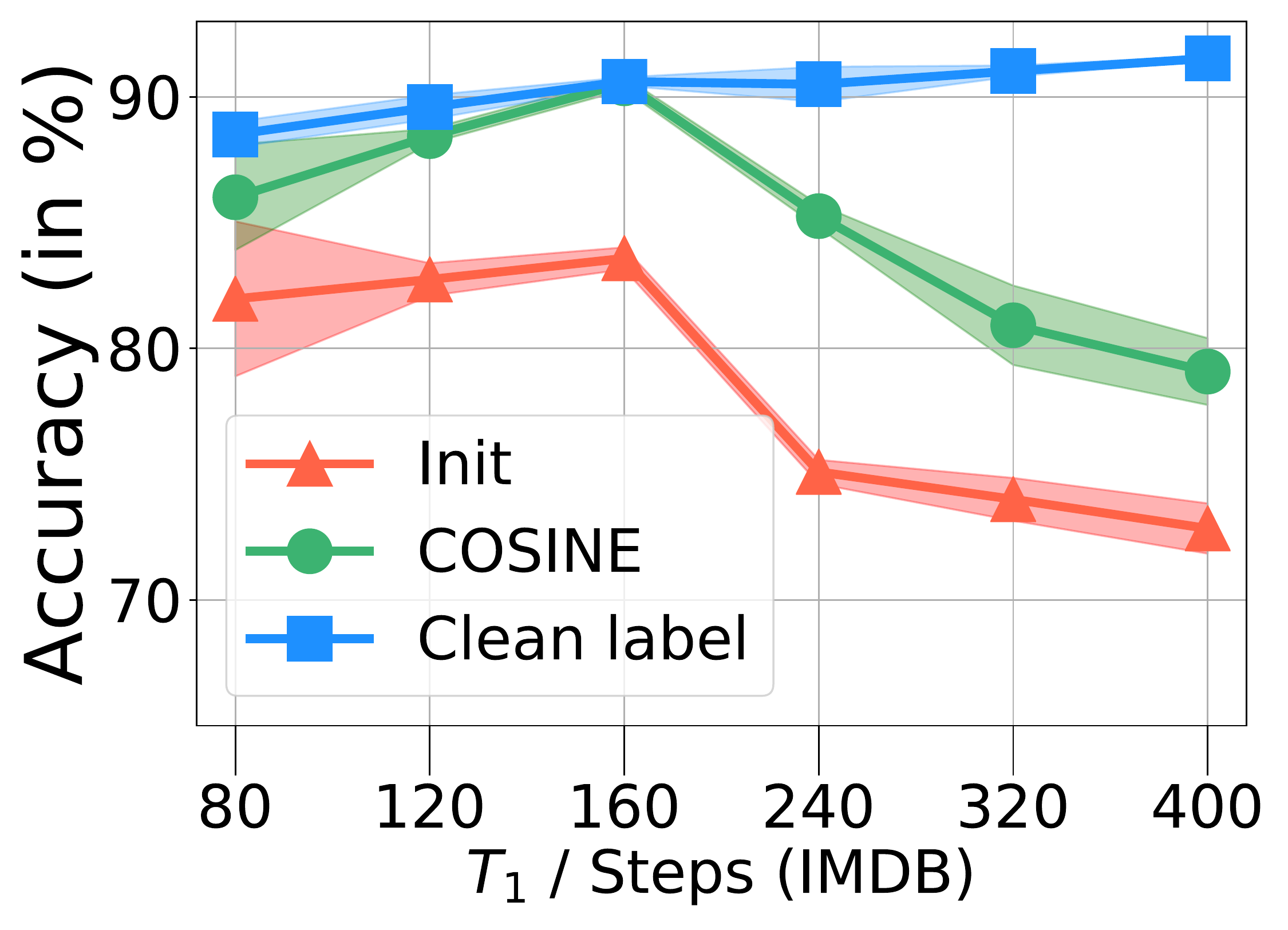}
            \label{fig:t1}
        }\hspace{-2.mm}
            \vspace{-1ex}
         \subfigure[\textit{Effect of $T_3$.}]{
            \includegraphics[width=0.32\textwidth]{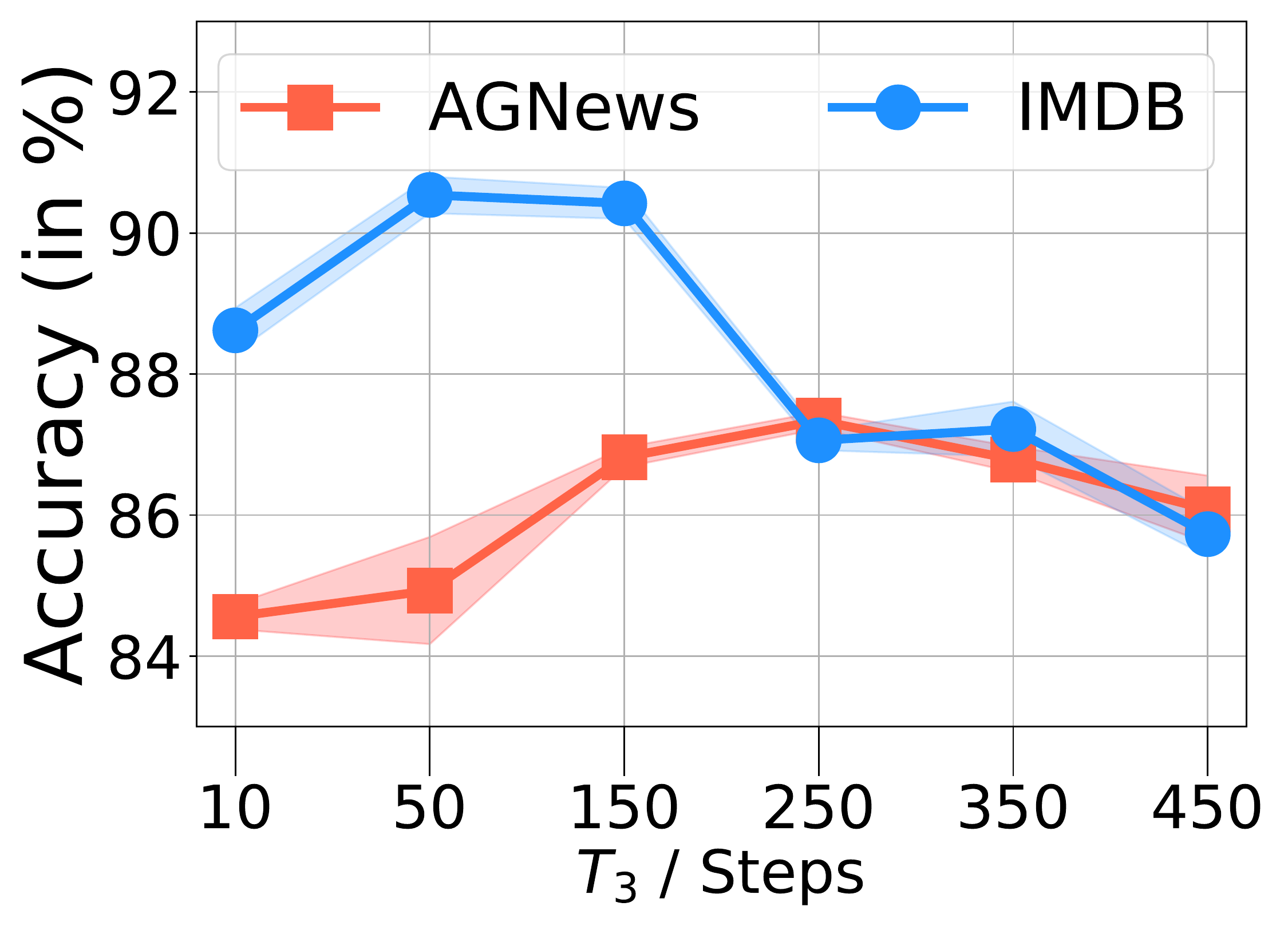}
            \label{fig:t3}
        }
        \vskip -0.05in
        \caption{\textit{Effects of different hyper-parameters.}}
        \label{fig:param}
    \end{minipage}%
    \begin{minipage}{.25\textwidth}
        \centering
        \includegraphics[width=0.99\textwidth]{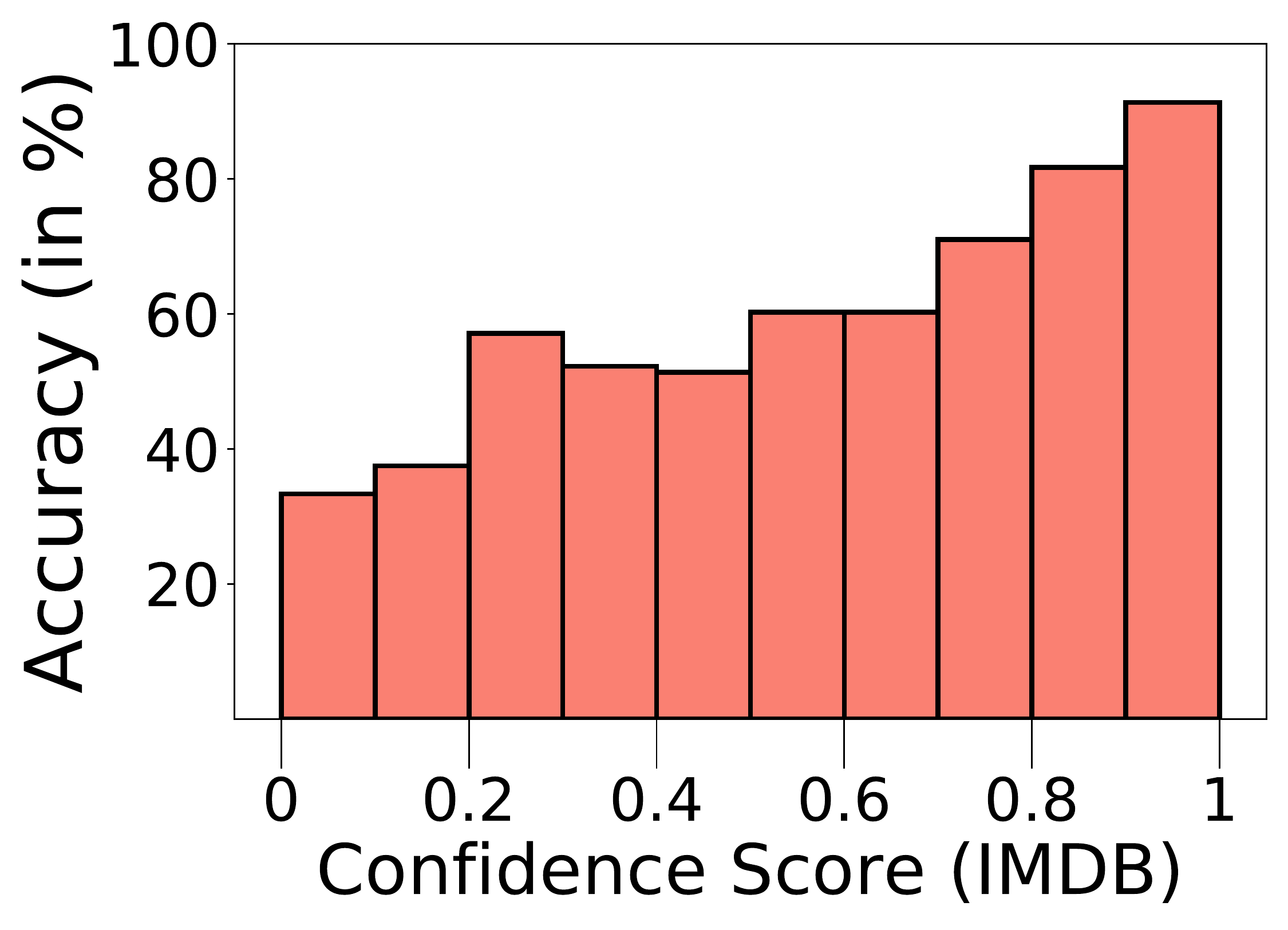}
        \vskip -0.05in
        \caption{\textit{Accuracy vs. Confidence score.}}
        \label{fig:confscore}
    \end{minipage}
    \vskip -0.2in
\end{figure*}


%% file: 5.6case.tex
\vspace{-0.1in}
\subsection{Case Study}
\vspace{-0.05in}

\begin{figure}[h]
    \vspace{-0.05in}
        \centering
        \hspace{-1.6mm}
        \subfigure[\textit{Embedding w/o $\cR_1$.}]{
            \includegraphics[width=0.231\textwidth]{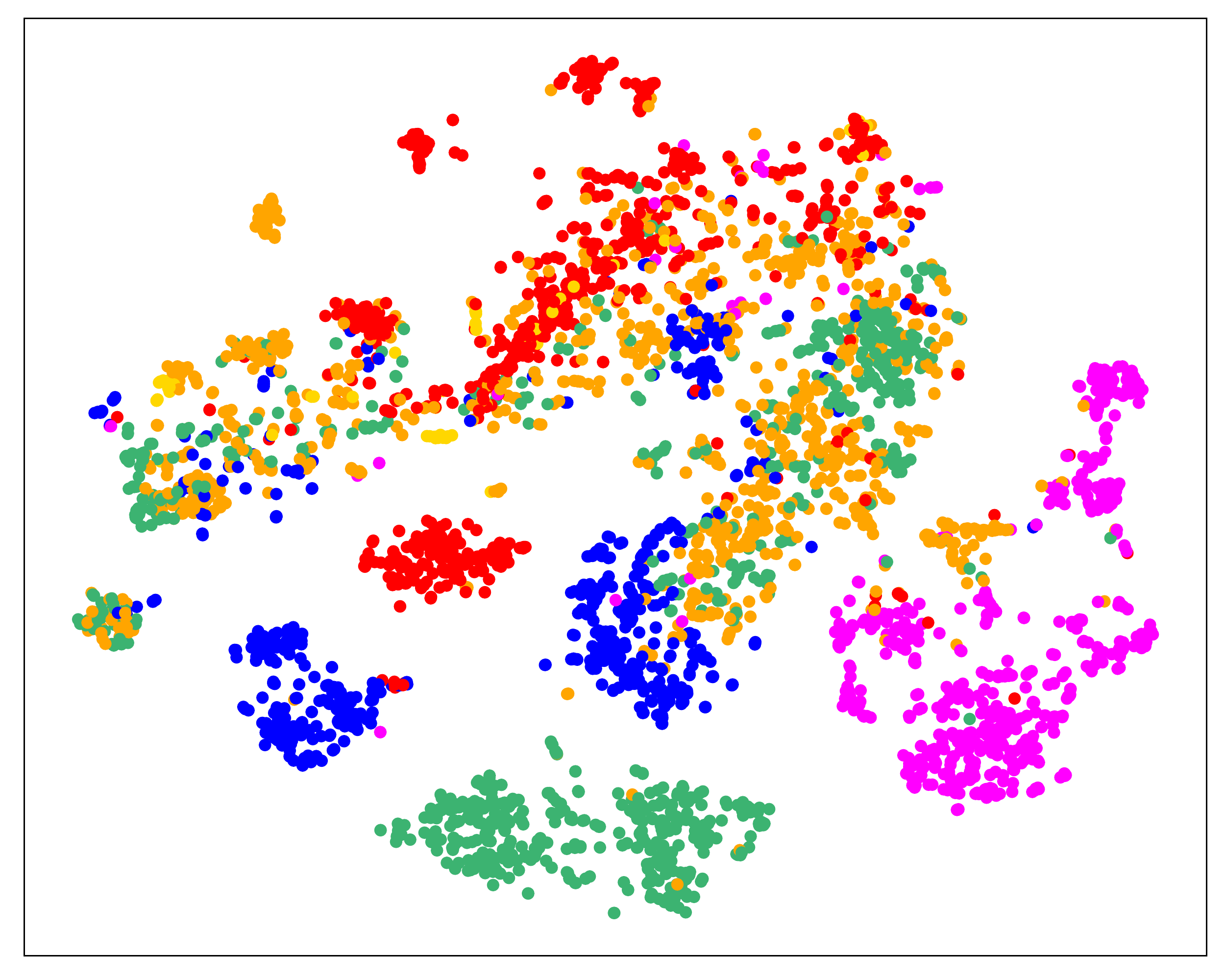}
            \label{fig:noco}
        }\hspace{-1.6mm}
        \subfigure[\textit{Embedding w/ $\cR_1$.}]{
            \includegraphics[width=0.231\textwidth]{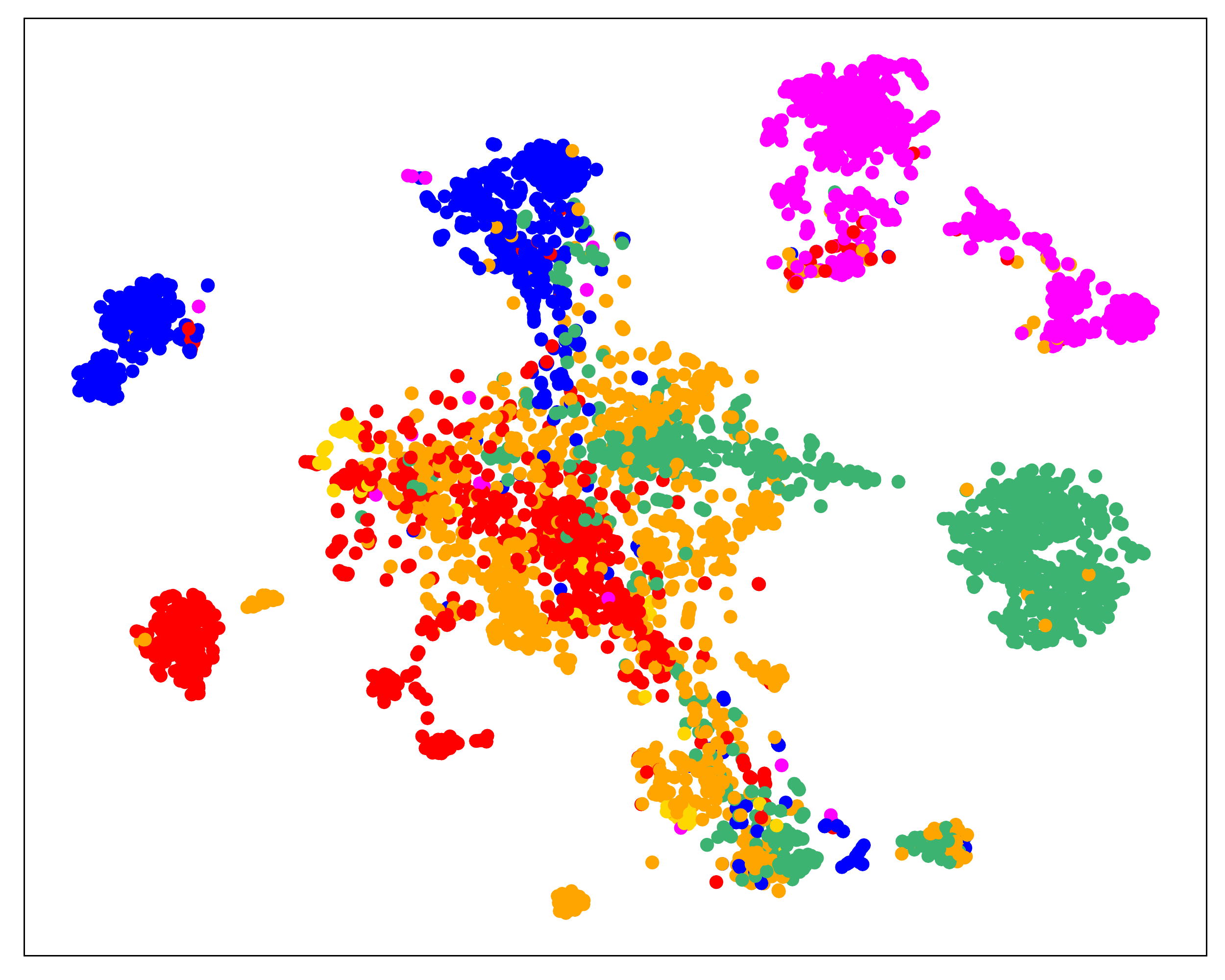}
            \label{fig:addco}
        }
        \hspace{-1.6mm}
        \vspace{-0.05in}
        \caption{\textit{t-SNE {\cite{tsne}} visualization on TREC. Each color denotes a different class.}}
        \label{fig:tsne}
    \vspace{-0.1in}
\end{figure}

\begin{figure}[t!]
  \vspace{-0.05in}
        \centering
        \hspace{-2.5mm}
        \subfigure[\textit{Results on Agnews.}]{
            \includegraphics[width=0.24\textwidth]{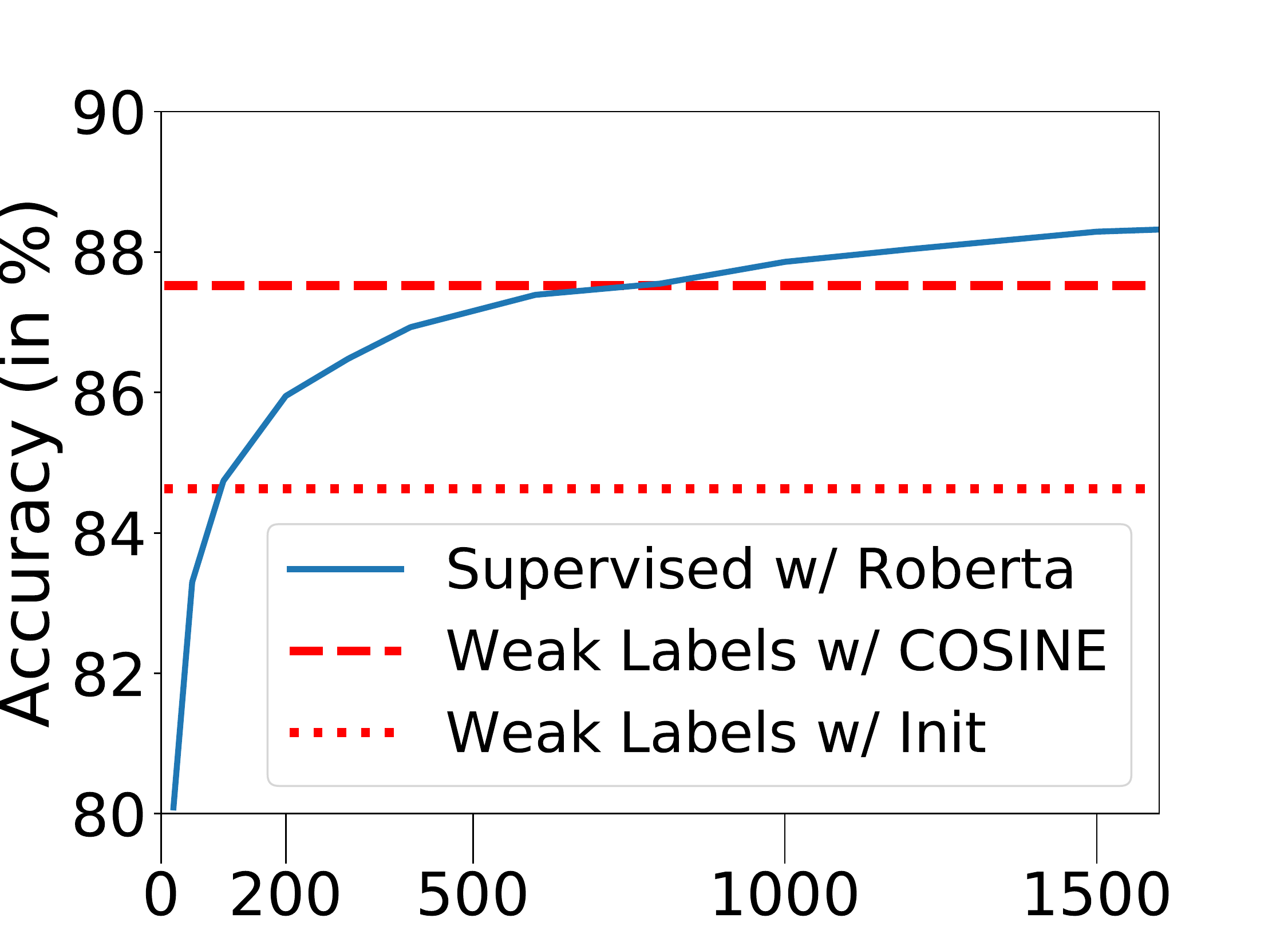}
            \label{fig:super_ag}
        }\hspace{-3.5mm}
        \subfigure[\textit{Results on MIT-R.}]{
            \includegraphics[width=0.24\textwidth]{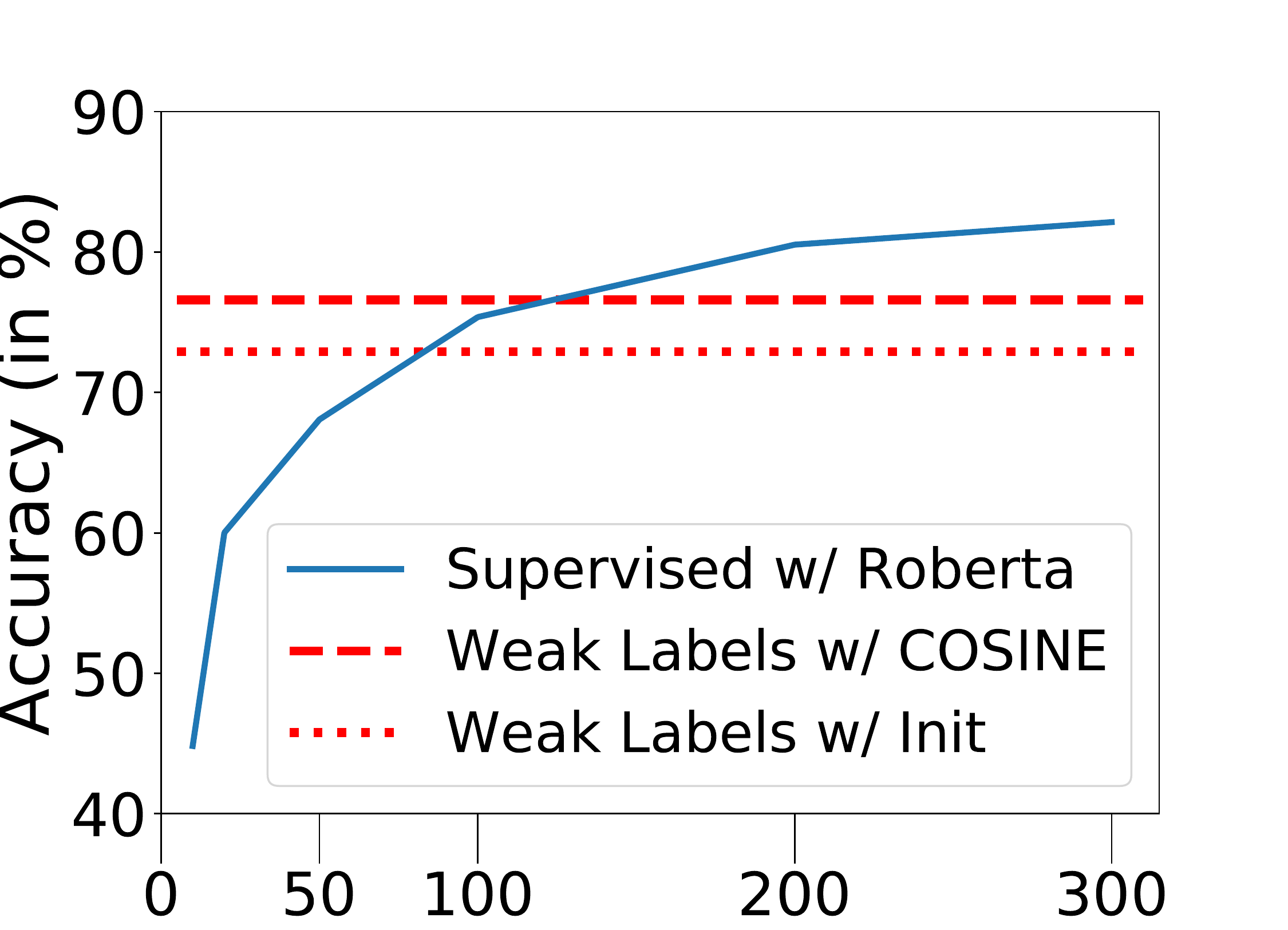}
            \label{fig:super_mit}
        }
        \hspace{-2.5mm}
        \vspace{-0.1in}
        \caption{\textit{Accuracy vs. Number of annotated labels.}}
        \label{fig:super}
    \vspace{-0.15in}
\end{figure}

\textbf{Error propagation mitigation and wrong-label correction.} Figure~\ref{fig:case} visualizes this process. Before training, the semantic rules make noisy predictions. After the initialization step, model predictions are less noisy but more biased, \eg, many samples are mis-labeled as ``\text{Amenity}''. These predictions are further refined by contrastive self-training.
The rightmost figure demonstrates \emph{wrong-label correction}. Samples are indicated by radii of the circle, and classification correctness is indicated by color, \ie, blue means correct and orange means incorrect. From inner to outer tori specify classification accuracy after the initialization stage, and the iteration 1,2,3. We can see that many incorrect predictions are corrected within three iterations. To illustrate: the right black dashed line means the corresponding sample is classified correctly after the first iteration, and the left dashed line indicates the case where the sample is mis-classified after the second iteration but corrected after the third. These results demonstrate that our model can correct wrong predictions via contrastive self-training.


\noindent\textbf{Better data representations.} We visualize sample embeddings in Fig.~\ref{fig:tsne}. By incorporating the contrastive regularizer $\cR_1$, our model learns more compact representations for data in the same class, \eg, the green class, and also extends the inter-class distances, \eg, the purple class is more separable from other classes in Fig.~\ref{fig:addco} than in Fig.~\ref{fig:noco}. 



\begin{figure}[t!]
  \centering
    \includegraphics[width=0.4\textwidth]{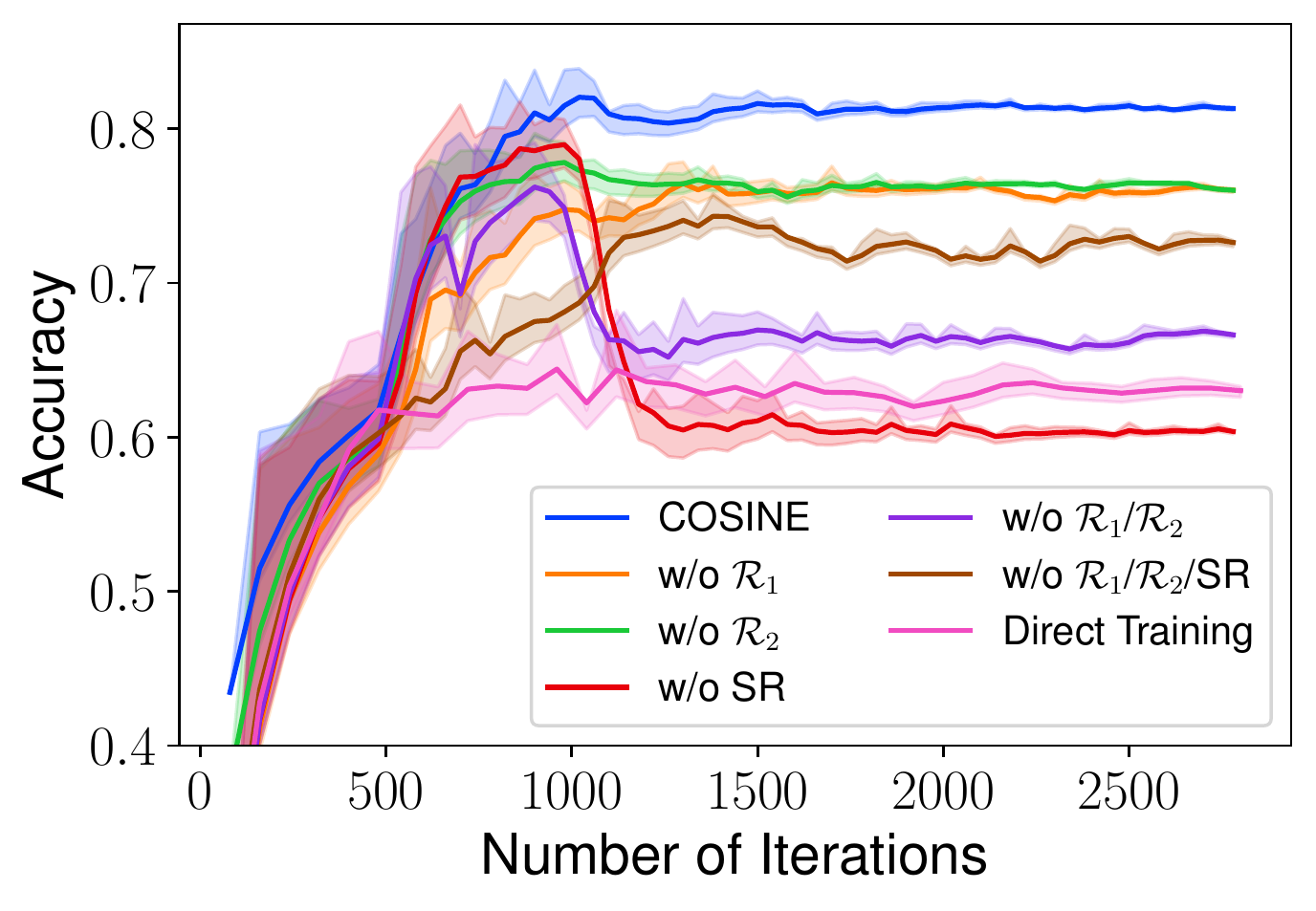}
    \vspace{-0.1in}
    \caption{\textit{Learning curves on TREC with different settings. Mean and variance are calculated over 3 runs.} }
  \label{fig:trec-ablation}
  \vspace{-0.05in}
\end{figure}

\noindent\textbf{Label efficiency.} Figure~\ref{fig:super} illustrates the number of clean labels needed for the supervised model to outperform \ours. On both of the datasets, the supervised model requires a significant amount of clean labels (around 750 for Agnews and 120 for MIT-R) to reach the level of performance as ours, whereas our method assumes no clean sample. 

\noindent\textbf{Higher Confidence Indicates Better Accuracy.} Figure~\ref{fig:confscore} demonstrates the relation between prediction confidence and prediction accuracy on IMDB.
We can see that in general, samples with higher prediction confidence yield higher prediction accuracy.
With our sample reweighting method, we gradually filter out low-confidence samples and assign higher weights for others, which effectively mitigates error propagation.

%% file: 5.2ablation.tex
\subsection{Ablation Study}
\label{sec:ablation}

\noindent \textbf{Components of {\ours}.} We inspect the importance of various components, including the contrastive regularizer $\cR_1$, the confidence regularizer $\cR_2$, and the sample reweighting (SR) method, and the soft labels. Table~\ref{tb:ablation_study} summarizes the results and Fig.~\ref{fig:trec-ablation} visualizes the learning curves. We remark that all the components jointly contribute to the model performance, and removing any of them hurts the classification accuracy. For example, sample reweighting is an effective tool to reduce error propagation, and removing it causes the model to eventually overfit to the label noise, \eg, the red bottom line in Fig.~\ref{fig:trec-ablation} illustrates that the classification accuracy increases and then drops rapidly.
On the other hand, replacing the soft pseudo-labels (Eq.~\ref{eq:reweilabel}) with the hard counterparts (Eq.~\ref{eq:hard}) \textbf{causes drops} in performance. This is because hard pseudo-labels lose prediction  confidence information. 

\begin{table}[h]
	\begin{small}
		\begin{center}
			\begin{tabular}{@{\hskip1pt}l@{\hskip2pt} |@{\hskip1.5pt} c @{\hskip1.5pt}|@{\hskip2pt} c @{\hskip2pt}|@{\hskip2pt} c @{\hskip2pt}|@{\hskip1.5pt} c@{\hskip1.5pt}|@{\hskip2pt} c@{\hskip2pt}}
				\toprule
				\textbf{Method} & \textbf{AGNews} & \textbf{IMDB} & \textbf{Yelp} & \textbf{MIT-R} & \textbf{TREC} \\
				\hline
				\Init & $84.63$ & $83.58$ & $81.76$ & $72.97$ &$66.50$\\
				\hline
				{\ours} & \textbf{87.52} & \textbf{90.54} & \textbf{95.97} & \textbf{76.61} & \textbf{82.59} \\
				~~w/o $\cR_1$ & $86.04$& $88.32$&$94.64$ & $74.11$ & $78.28$ \\
				~~w/o $\cR_2$ & $85.91$  & $89.32$ &$93.96$ & $75.21$ & $77.11$  \\
				~~w/o SR & $86.72$  & $87.10$ & $93.08$ & $74.29$ & $79.77$  \\
				~~w/o $\cR_1$/$\cR_2$ & $86.33$ & $84.44$& $92.34$& $73.67$ & $76.95$  \\
				~~w/o $\cR_1$/$\cR_2$/SR & $86.61$ & $83.98$ & $82.57$ & $73.59$ & $74.96$ \\
				~~w/o Soft Label  &$86.07$ & $89.72$ & $93.73$ & $73.05$ & $71.91$ \\
				\bottomrule
			\end{tabular}
		\end{center}
	\end{small}
	\vspace{-0.1in}
    \caption{\textit{Effects of different components. Due to space limit we only show results for 5 representative datasets.}}
    \vspace{-0.1in}
    \label{tb:ablation_study}
\end{table}

\noindent \textbf{Hyper-parameters of {\ours}.} In Fig.~\ref{fig:param}, we examine the effects of different hyper-parameters, including the confidence threshold $\xi$ (Eq.~\ref{eq:C}), the stopping time $T_1$ in the initialization step, and the update period $T_3$ for pseudo-labels.  
From Fig.~\ref{fig:xi}, we can see that setting the confidence threshold too big hurts model performance, which is because an over-conservative selection strategy can result in insufficient number of training data. 
The stopping time $T_1$ has drastic effects on the model. This is because fine-tuning {\ours} with weak labels for excessive steps causes the model to unavoidably overfit to the label noise, such that the contrastive self-training procedure cannot correct the error.
Also, with the increment of $T_3$, the update period of pseudo-labels, model performance first increases and then decreases. This is because if we update pseudo-labels too frequently, the contrastive self-training procedure cannot fully suppress the label noise, and if the updates are too infrequent, the pseudo-labels cannot capture the updated information well. 

%% file: 5Relatedwork.tex
\section{Related Works}
\vspace{-0.1in}


\noindent \textbf{Fine-tuning Pre-trained Language Models.}
To improve the model's generalization power during fine-tuning stage, several methods are proposed~\cite{peters2019tune,dodge2020fine,Zhu2020FreeLB,smart,ensemble,kong2020calibrated,zhao2020masking,gunel2021supervised,zhang2021revisiting,aghajanyan2021better,wang2021infobert}, 
However,  most of these methods focus on fully-supervised setting and rely heavily on large amounts of \emph{clean labels}, which are not always available. To address this issue, we propose a contrastive self-training framework that fine-tunes pre-trained models with only weak labels. Compared with the existing fine-tuning approaches~\cite{ensemble,Zhu2020FreeLB,smart}, our model effectively reduce the label noise, which achieves better performance on various NLP tasks with weak supervision.



\noindent \textbf{Learning From Weak Supervision.} In weakly-supervised learning, the training data are usually noisy and incomplete. Existing methods aim to denoise the sample labels or the labeling functions by, for example, aggregating multiple weak supervisions~\cite{ratner2019snorkel,lison2020named,ren2020denoise}, using clean samples~\cite{Awasthi2020Learning}, and leveraging contextual information~\cite{mekala2020contextual}. However, most of them can only use specific type of weak supervision on specific task, \eg, keywords for text classification~\cite{meng2020weakly,mekala2020contextual}, and they require prior knowledge on weak supervision sources~\cite{Awasthi2020Learning,lison2020named,ren2020denoise}, which somehow limits the scope of their applications.
Our work is orthogonal to them since we do not denoise the labeling functions directly. 
Instead, we adopt contrastive self-training to leverage the power of  pre-trained language models for denoising, which
is \emph{task-agnostic} and applicable to various NLP tasks with minimal
additional efforts.

%% file: 6discussion.tex
\vskip -0.1in
\section{Discussions}
\vskip -0.1in


\noindent \textbf{Adaptation of LMs to Different Domains.} When fine-tuning LMs on data from different domains, we can first continue pre-training on in-domain text data for better adaptation~\cite{gururangan2020dont}. For some rare domains where BERT trained on general domains is not optimal, we can use LMs pretrained on those specific domains (\eg BioBERT~\cite{lee2020biobert}, SciBERT~\cite{beltagy2019scibert}) to tackle this issue.

\noindent \textbf{Scalability of Weak Supervision.} {\ours} can be applied to tasks with a large number of classes. This is because rules can be automatically generated beyond hand-crafting. For example, we can use label names/descriptions as weak supervision signals~\cite{meng2020weakly}. Such signals are easy to obtain and do not require hand-crafted rules. Once weak supervision is provided, we can create weak labels to further apply {\ours}.

\noindent \textbf{Flexibility.} {\ours} can handle tasks and weak supervision sources beyond our conducted experiments. For example, other than semantic rules, crowd-sourcing can be another weak supervision source to generate pseudo-labels~\cite{wang2013perspectives}. Moreover, we only conduct experiments on several representative tasks, but our framework can be applied to other tasks as well, e.g., named-entity recognition (token classification) and reading comprehension (sentence pair classification).

%% file: 7conclusion.tex
\vskip -0.3in
\section{Conclusion}
\vskip -0.1in

In this paper, we propose a contrastive regularized self-training framework, {\ours}, for fine-tuning pre-trained language models with weak supervision. Our framework can learn better data representations to ease the classification task, and also efficiently reduce label noise propagation by confidence-based reweighting and regularization. We conduct experiments on various classification tasks, including sequence classification, token classification, and sentence pair classification, and the results demonstrate the efficacy of our model.

\vskip -0.1in
\section*{Broader Impact}
\vskip -0.1in

{\ours} is a general framework that tackled the label scarcity issue via 
\textbf{combining neural nets with
weak supervision}. The weak supervision provides a simple but flexible language to encode the domain knowledge and capture the correlations between features and labels. When combined with unlabeled data, our framework can largely tackle the label scarcity bottleneck for training DNNs, enabling them to be applied for downstream NLP classification tasks in a label efficient manner.

{\ours} neither introduces any social/ethical bias to the model nor amplify any bias in the data. In all the experiments, we use publicly available data, and we build our algorithms using public code bases. We do not foresee any direct social consequences or ethical issues.

%% file: appendix_upload.tex
\clearpage
\section{Weak Supervision Details}
\label{app:dataset}










\label{app:LFs}
\input{supple-rules}


\section{Baseline Settings} \label{app:baseline-settings}
We implement Self-ensemble, FreeLB, Mixup and UST based on their original paper. 
For other baselines, we use their official release: \\
\noindent $\diamond$ WeSTClass~\cite{meng2018weakly}:~\url{https://github.com/yumeng5/WeSTClass}. \\
\noindent $\diamond$ RoBERTa~\cite{liu2019roberta}:~\url{https://github.com/huggingface/transformers}. \\
\noindent $\diamond$ SMART~\cite{smart}:~\url{https://github.com/namisan/mt-dnn}. \\
\noindent $\diamond$ Snorkel~\cite{ratner2019snorkel}:~\url{https://www.snorkel.org/}. \\
\noindent $\diamond$ ImplyLoss~\cite{Awasthi2020Learning}:~\url{https://github.com/awasthiabhijeet/Learning-From-Rules}. \\
\noindent $\diamond$ Denoise~\cite{ren2020denoise}:~\url{https://github.com/weakrules/Denoise-multi-weak-sources}.

\section{Details on Experiment Setups}
\label{appendix:exp}

\subsection{Computing Infrastructure}
\textbf{System}: Ubuntu 18.04.3 LTS; Python 3.7; Pytorch 1.2. \textbf{CPU}: Intel(R) Core(TM) i7-5930K CPU @ 3.50GHz. \textbf{GPU}: GeForce GTX TITAN X. \\
\vspace{-3mm}
\subsection{Hyper-parameters}
We use AdamW \cite{loshchilov2018adamw} as the optimizer, and the learning rate is chosen from 
$1\times 10^{-5}, 2\times 10^{-5}, 3 \times 10^{-5}\}$. A linear learning rate decay schedule with warm-up $0.1$ is used, and the number of training epochs is $5$.

Hyper-parameters are shown in Table~\ref{tab:hyperparameter}. We use a grid search to find the optimal setting for each task. Specifically, we search $T_1$ from $10$ to $2000$, $T_2$ from $1000$ to $5000$, $T_3$ from $10$ to $500$,  $\xi$ from $0$ to $1$, and $\lambda$ from $0$ to $0.5$. All results are reported as the average over three runs.

\begin{table*}[htb!]
	\begin{center}
		\begin{tabular}{c|c|c|c|c|c|c|c}
			\toprule 
			\bf Hyper-parameter &\bf AGNews& \bf  IMDB & \bf Yelp  & \bf MIT-R & \bf TREC  & \bf Chemprot & \bf WiC\\ \midrule 
			Dropout Ratio & \multicolumn{7}{@{\hskip1pt}c@{\hskip1pt}}{0.1}  \\ \hline
			 Maximum Tokens  & 128 & 256 & 512 & 64 & 64 & 400 & 256 \\ \hline 
			 Batch Size  & 32 & 16 & 16 & 64 & 16 & 24 & 32 \\ \hline
			 Weight Decay & \multicolumn{7}{@{\hskip1pt}c@{\hskip1pt}}{$10^{-4}$}  \\ \hline
			 Learning Rate & $10^{-5}$ & $10^{-5}$ & $10^{-5}$ & $10^{-5}$  & $10^{-5}$ & $10^{-5}$ & $10^{-5}$\\ \hline
			 $T_1$ & 160 & 160 & 200 & 150 & 500 & 400 & 1700\\ \hline
             $T_2$ &  3000 & 2500 & 2500 & 1000 & 2500 & 1000 & 3000 \\ \hline
             $T_3$ &  250 & 50& 100& 15  & 30 & 15 & 80 \\ \hline
			 $\xi$ & 0.6 &0.7 & 0.7 & 0.2 & 0.3 & 0.7 & 0.7\\ \hline
             $\lambda$ & 0.1 & 0.05 & 0.05 & 0.1 & 0.05 & 0.05 & 0.05\\
			 \bottomrule
		\end{tabular}
	\end{center}
	\vspace{-4mm}
	\caption{Hyper-parameter configurations. Note that we only keep certain number of tokens.}
	\label{tab:hyperparameter}
\end{table*}

\subsection{Number of Parameters}
{\ours} and most of the baselines (RoBERTa-WL / RoBERTa-CL / SMART / WeSTClass / Self-Ensemble / FreeLB / Mixup / UST) are built on the RoBERTa-base model with about 125M parameters.
Snorkel is a generative model with only a few parameters.
ImplyLoss and Denoise freezes the embedding and has less than 1M parameters.
However, these models cannot achieve satisfactory performance in our experiments.

\begin{table*}[h]
	\begin{center}
		\begin{tabular}{c|c|c|c|c|c|c}
			\toprule 
			\bf Distance $d$ &\multicolumn{3}{c|}{Euclidean}  & \multicolumn{3}{c}{Cos} \\ \hline 
			\bf Similarity $W$ & Hard  &   KL-based  &  L2-based     & Hard  &   KL-based &  L2-based \\ \midrule 
			 AGNews & \textbf{87.52} & 86.44&86.72 & 87.34 &86.98&86.55  \\ \hline 
			 MIT-R & 76.61 &76.68&76.49 & 76.55&\textbf{76.76}&76.58 \\  \bottomrule
		\end{tabular}
	\end{center}
	\vspace{-4mm}
	\caption{Performance of {\ours} under different settings.}
	\label{tab:dw}
\end{table*}


\section{Early Stopping and Earlier Stopping}
\label{appendix:early}
Our model adopts the earlier stopping strategy during the initialization stage. Here we use ``earlier stopping'' to differentiate from ``early stopping'', which is standard in fine-tuning algorithms.
Early stopping refers to the technique where we stop training when the evaluation score drops. Earlier stopping is self-explanatory, namely we fine-tune the pre-trained LMs with only a few steps, even before the evaluation score starts dropping.
This technique can efficiently prevent the model from overfitting. For example, as Figure~\ref{fig:t1} illustrates, on IMDB dataset, our model overfits after 240 iterations of initialization with weak labels. In contrast, the model achieves good performance even after 400 iterations of fine-tuning when using clean labels. This verifies the necessity of earlier stopping.

\section{Comparison of Distance Measures in Contrastive Learning}
\label{appendix:distance}

The contrastive regularizer $\cR_1(\theta; \tilde{\by})$ is related to two designs: the sample distance metric $d_{ij}$ and the sample similarity measure $W_{ij}$. In our implementation, we use the scaled Euclidean distance as the default for $d_{ij}$ and Eq.~\ref{eq:edgeweight} as the default for $W_{ij}$\footnote{To accelerate contrastive learning, we adopt the doubly stochastic sampling approximation to reduce the computational cost. Specifically, the high confidence samples $\cC$ in each batch $\cB$ yield $\cO(|\cC|^2)$ sample pairs, and we sample $|\cC|$ pairs from them.}.
Here we discuss other designs.

\subsection{Sample distance metric $d$}
Given the encoded vectorized representations $\bv_i$ and $\bv_j$ for samples $i$ and $j$, we consider two distance metrics as follows.

\noindent \textbf{Scaled Euclidean distance (Euclidean)}: We calculate the distance between $\bv_i$ and $\bv_j$ as 
\begin{equation}
d_{i j}=\frac{1}{d}\left\|\boldsymbol{v}_{i}-\boldsymbol{v}_{j}\right\|_{2}^{2}.
\end{equation}

\noindent \textbf{Cosine distance (Cos)}\footnote{We use Cos to distinguish from our model name {\ours}.}: 
Besides the scaled Euclidean distance, cosine distance is another widely-used distance metric:
\begin{equation}
d_{i j}=1 - \text{cos}\left(\bv_i, \bv_j\right) = 1 - \frac{\|\bv_{i}\cdot \bv_{j}\|}{\|\bv_{i}\| \|\bv_{j}\|}.
\end{equation}

\subsection{Sample similarity measures $W$}
Given the soft pseudo-labels $\tilde{\by}_i$ and $\tilde{\by}_j$ for samples $i$ and $j$, the following are some designs for $W_{ij}$. In all of the cases, $W_{ij}$ is scaled into range $[0, 1]$ (we set $\gamma=1$ in Eq.~\ref{eq:ell} for the hard similarity).

\noindent \textbf{Hard Similarity}: The hard similarity between two samples is calculated as
\begin{equation}
W_{i j}=\left\{\begin{array}{ll}
1, & \text { if } \underset{k \in \cY}{\text{argmax}}[\tilde{\by}_i]_k = \underset{k \in \cY}{\text{argmax}}[\tilde{\by}_j]_k, \\
0, & \text { otherwise. }
\end{array}\right.
\end{equation}
This is called a ``hard'' similarity because we obtain a binary label, \ie, we say two samples are similar if their corresponding hard pseudo-labels are the same, otherwise we say they are dissimilar.

\noindent \textbf{Soft KL-based Similarity}: We calculate the similarity based on KL distance as follows.
\begin{equation} 
W_{i j}= \exp\left( -\frac{\beta}{2} \Big(\cD_{KL}(\tilde{\by}_i \| \tilde{\by}_j) + \cD_{KL}(\tilde{\by}_j \| \tilde{\by}_i)\Big) \right),
\end{equation}
where $\beta$ is a scaling factor, and we set $\beta = 10$ by default.

\noindent \textbf{Soft L2-based Similarity}: We calculate the similarity based on L2 distance as follows.
\begin{equation} 
W_{i j}= 1- \frac{1}{2}||\tilde{\by}_i - \tilde{\by}_j||_2^2,
\end{equation}

\subsection{{\ours} under different $d$ and $W$.}
We show the performance of {\ours} with different choices of $d$ and $W$ on Agnews and MIT-R in Table \ref{tab:dw}. We can see that {\ours} is robust to these choices.
In our experiments, we use the scaled euclidean distance and the hard similarity by default.








%% file: supple-rules.tex







{\ours} does not require any human annotated examples in the training process, and it only needs weak supervision sources such as keywords and semantic rules. According to some studies in existing works \citet{Awasthi2020Learning, nero}, such weak supervisions are cheap to obtain and are much efficient than collecting clean labels. In this way, we can obtain significantly more labeled examples using these weak supervision sources than human labor.

There are two types of semantic rules that we apply as weak supervisions:
\begin{enumerate}
    \item[$\diamond$] \textit{Keyword Rule}: \texttt{HAS(x, L) $\rightarrow$ C}. If $x$ matches one of the words in the list $L$, we label it as $C$.
    \item[$\diamond$] \textit{Pattern Rule}: \texttt{MATCH(x, R) $\rightarrow$ C}. If $x$ matches the regular expression $R$, we label it as $C$.
\end{enumerate}
In addition to the keyword rule and the pattern rule, we can also use third-party tools to obtain weak labels. These tools (\eg TextBlob\footnote{\url{https://textblob.readthedocs.io/en/dev/index.html.}}) are available online and can be obtained cheaply, but their prediction is not accurate enough (when directly use this tool to predict label for all training samples, the accuracy on Yelp dataset is around 60\%). \\
We now introduce the semantic rules on each dataset:
\begin{enumerate}
    \item[$\diamond$] \textit{AGNews, IMDB, Yelp}: We use the rule in~\citet{ren2020denoise}. Please refer to the original paper for detailed information on rules.
    \item[$\diamond$] \textit{MIT-R, TREC}:  We use the rule in~\citet{Awasthi2020Learning}. Please refer to the original paper for detailed information on rules.
    \item[$\diamond$] \textit{ChemProt}: There are 26 rules. We show part of the rules in Table \ref{tab:chem}.
    \item[$\diamond$] \textit{WiC}: Each sense of each word in WordNet has example sentences. For each sentence in the WiC dataset and its corresponding keyword, we collect the example sentences of that word from WordNet. Then for a pair of sentences, the corresponding weak label is ``True'' if their definitions are the same, otherwise the weak label is ``False''.
\end{enumerate}

\begin{table*}[htb!]\small
    \centering
    \begin{tabular}
    {p{150pt} | p{250pt}}
    \toprule
         \textbf{Rule} & \textbf{Example}\\
         \midrule
         \texttt{HAS (x, [amino acid,mutant, mutat, replace] ) $\rightarrow$ part\_of } 
         & A major part of this processing requires endoproteolytic cleavage at specific pairs of basic [CHEMICAL]amino acid[CHEMICAL] residues, an event necessary for the maturation of a variety of important biologically active proteins, such as insulin and [GENE]nerve growth factor[GENE].\\ \hline
         
         \texttt{HAS (x, [bind, interact, affinit] ) $\rightarrow$ regulator} &
         The interaction of [CHEMICAL]naloxone estrone azine[CHEMICAL] (N-EH) with various [GENE]opioid receptor[GENE] types was studied in vitro.\\ \hline
         
         \texttt{HAS (x, [activat, increas, induc, stimulat, upregulat] ) $\rightarrow$ upregulator/activator }
         & The results of this study suggest that [CHEMICAL]noradrenaline[CHEMICAL] predominantly, but not exclusively, mediates contraction of rat aorta through the activation of an [GENE]alphalD-adrenoceptor[GENE]. \\ \hline
        
        \texttt{HAS (x, [downregulat, inhibit, reduc, decreas] ) $\rightarrow$ downregulator/inhibitor }
        & These results suggest that [CHEMICAL]prostacyclin[CHEMICAL] may play a role in downregulating [GENE]tissue factor[GENE] expression in monocytes, at least in part via elevation of intracellular levels of cyclic AMP.\\ \hline
        
        \texttt{HAS (x, [ agoni, tagoni]* ) $\rightarrow$ agonist}  * (note the leading whitespace in both cases)
         & Alprenolol and BAAM also caused surmountable antagonism of [CHEMICAL]isoprenaline[CHEMICAL] responses, and this [GENE]beta 1-adrenoceptor[GENE] antagonism was slowly reversible. \\ \hline
         
         \texttt{HAS (x, [antagon] ) $\rightarrow$ antagonist }
         & It is concluded that [CHEMICAL]labetalol[CHEMICAL] and dilevalol are [GENE]beta 1-adrenoceptor[GENE] selective antagonists. \\ \hline
         
         \texttt{HAS (x, [modulat, allosteric] ) $\rightarrow$ modulator}
         & [CHEMICAL]Hydrogen sulfide[CHEMICAL] as an allosteric modulator of [GENE]ATP-sensitive potassium channels[GENE] in colonic inflammation. \\ \hline
         
         \texttt{HAS (x, [cofactor] ) $\rightarrow$ cofactor}
         & The activation appears to be due to an increase of [GENE]GAD[GENE] affinity for its cofactor, [CHEMICAL]pyridoxal phosphate[CHEMICAL] (PLP). \\ \hline
         
         \texttt{HAS (x, [substrate, catalyz, transport, produc, conver] ) $\rightarrow$ substrate/product}
         & Kinetic constants of the mutant [GENE]CrAT[GENE] showed modification in favor of longer [CHEMICAL]acyl-CoAs[CHEMICAL] as substrates. \\ \hline
         
         \texttt{HAS (x, [not] ) $\rightarrow$ not}
          & [CHEMICAL]Nicotine[CHEMICAL] does not account for the CSE stimulation of [GENE]VEGF[GENE] in HFL-1. \\
    \bottomrule
    \end{tabular}
    \caption{Examples of semantic rules on Chemprot.}
    \label{tab:chem}
\end{table*}
